\documentclass{article}

\usepackage{arxiv}
\usepackage{booktabs}                 
\usepackage{tabularx}                 
\usepackage{array}                    
\usepackage[table]{xcolor}            
\usepackage{ragged2e}                 
\usepackage{makecell}                 
\usepackage{algorithm}
\usepackage{algpseudocode}
\usepackage[utf8]{inputenc}           
\usepackage{amsmath}
\usepackage{amssymb}
\usepackage[T1]{fontenc}              
\usepackage{hyperref}                 
\usepackage{url}                      
\usepackage{amsfonts}                 
\usepackage{nicefrac}                 
\usepackage{microtype}                
\usepackage{cleveref}                 
\usepackage{graphicx}
\usepackage{natbib}
\usepackage{doi}

\newcolumntype{L}[1]{>{\RaggedRight\arraybackslash}p{#1}}
\newcolumntype{Y}{>{\RaggedRight\arraybackslash}X}
\floatname{algorithm}{Algorithm}

\title{A Cascaded Framework with a Cross-Attention-Guided Mamba-CNN Model for Pre- and Post-Treatment Glioma Segmentation}

\newif\ifuniqueAffiliation
\uniqueAffiliationtrue

\ifuniqueAffiliation 
\author{
	\href{}{Siavash Taremi}\\
	School of Medicine\\
	Zanjan University of Medical Sciences\\
	Zanjan, Iran \\
	\And
		\href{}{Mahmoud Heidari}\\
	School of Medicine\\
	Zanjan University of Medical Sciences\\
	Zanjan, Iran \\
	\And
	 \href{https://orcid.org/0009-0008-0210-6868}{\includegraphics[scale=0.06]{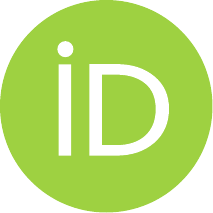}\hspace{1mm}Seyed Mohammad Hossein Hashemi}\\
	Department of CS \& IT\\
	Institute for Advanced Studies in Basic Sciences\\
	Zanjan, Iran \\
	\And
	\href{https://orcid.org/0000-0002-2934-2675}{\includegraphics[scale=0.06]{orcid.pdf}\hspace{1mm}Mohsen Hooshmand}\thanks{Corresponding author: mohsen.hooshmand@iasbs.ac.ir} \\
	Department of CS \& IT\\
	Institute for Advanced Studies in Basic Sciences\\
	Zanjan, Iran \\
}

\renewcommand{\shorttitle}{\textit{arXiv} Template}

\hypersetup{
pdftitle={A template for the arxiv style},
pdfsubject={q-bio.NC, q-bio.QM},
pdfauthor={David S.~Hippocampus, Elias D.~Striatum},
pdfkeywords={First keyword, Second keyword, More},
}

\begin{document}
\maketitle

\begin{abstract}
	Accurate glioma segmentation is vital for diagnosis and follow-up, but manual segmentation is time-consuming, subjective, and varies across interpreters. Gliomas also have heterogeneous structures, with variations in size, shape, and location. Thus, deep learning-based segmentation methods have emerged as a potential solution. Convolutional neural networks (CNNs) excel at capturing local features but may struggle to model long-range dependencies, while Vision Transformers (ViTs) capture global context efficiently but require higher computational complexity. Moreover, most previous studies have focused on pre-treatment gliomas, while post-treatment cases with resection cavities (RCs) remain comparatively less explored. This study proposes a Cross-Attention-Guided Mamba-CNN (CMC-Seg) model within a cascaded framework for pre- and post-treatment diffuse glioma segmentation, aiming to balance accuracy and computational cost. Experiments were conducted on the BraTS 2025 dataset, which includes multi-parametric magnetic resonance imaging (mpMRI) of patients with diffuse gliomas. After preprocessing, a coarse segmentation model was employed to extract tumor-centered image patches, which were subsequently fed to CMC-Seg for fine-grained segmentation. Comprehensive evaluations were conducted against 3D U-Net, SegResNet, SwinUNETR, UNETR, nnFormer, and SegMamba. On the test set, CMC-Seg achieved the highest mean Dice score (81.20\%) and the non-enhancing tumor core (NETC) Dice score (77.09\%). CMC-Seg also demonstrated a Dice score of 78.63\% and the lowest 95\% Hausdorff distance (HD95) score (4.39 mm) for the RC. These results were comparable to the CNN and Mamba baselines and significantly better than the Transformer-based baselines (P < 0.05). In addition, CMC-Seg required the fewest inference GFLOPs (381.49).  Ablation experiments indicated that multi-direction scanning and cross-attention improved model performance, particularly for the RC class (P < 0.05). Finally, the Grad-CAM explainability experiments showed that CMC-Seg focuses on diagnostically relevant tumor regions.
\end{abstract}


\section{Introduction}
Gliomas are intra-axial tumors of the central nervous system (CNS) that arise from glial cells and represent the most common primary CNS tumor. In the United States, glioma incidence is approximately 5.83 per 100,000 people annually (\cite{price2026cbtrus}). Also, gliomas represent approximately 30\% of all primary brain tumors and account for around 80\% of malignant brain tumors, making them the leading cause of death among patients with primary brain tumors (\cite{weller2015glioma}). Early detection of glioma, accurate delineation of tumor burden before surgery, and assessment of tumor response during and after treatment are essential. Magnetic resonance imaging (MRI) is the diagnostic gold standard for detecting brain tumors and for patient follow-up, typically including T1-weighted (T1w) sequences before and after gadolinium contrast enhancement (T1c), T2-weighted (T2w), and T2-weighted fluid-attenuated inversion recovery (FLAIR) sequences. Each modality offers distinct yet complementary information. For example, T2w and FLAIR highlight the edematous region surrounding the tumor, while T1w and T1c emphasize the tumor itself. By integrating these MRI sequences, radiologists and clinicians can develop a comprehensive view of a brain tumor, including its location, size, morphology, vascularity, and relationship to surrounding structures (\cite{weller2021eano}, \cite{zhou2019review}). 

Tumor segmentation is a crucial step in medical image analysis, assisting early diagnosis, treatment planning, and disease monitoring. The main imaging technique for segmenting glioma is MRI, which enables notable soft tissue clarity and provides a combination of several imaging sequences that emphasize various tissue properties (\cite{amri2025automatic}).  However, the complex nature of glioma, such as variations in size, shape, and location across various MRI sequences, causes significant challenges. Manually delineating each slice on MRI sequences by radiologists poses several limitations, such as being time-intensive, subjectivity, and variability in results among different interpreters (\cite{van2021performance}). Computer-aided diagnosis (CAD) systems aim to reduce these limitations by improving both diagnostic precision and workflow efficiency. However, classical CAD for segmentation, such as thresholding, region-based, edge-based, clustering-based, and traditional machine learning models, are usually not well suited for heterogeneous tumor structures and modality-specific variations (\cite{magadza2021deep}, \cite{zhang2021overview}).

In recent years, with the development of deep learning models, particularly Convolutional Neural Networks (CNNs), Artificial Intelligence (AI)-based methods have gained significant attention for medical image segmentation(\cite{ghadimi2025deep}). Most segmentation models rely on encoder-decoder architectures, with the U-Net model as a prime example. In the U-Net (\cite{ronneberger2015u}) model design, the encoder progressively extracts hierarchical, low-dimensional feature representations from the input images, and the decoder converts these learned features back into pixel-wise segmentation maps, using skip connections to preserve spatial information. UNet++ (\cite{zhou2018unet++}) uses nested and dense skip connections to reduce the semantic gap between encoder and decoder feature maps, guided by deep supervision. 

Although CNNs improve segmentation accuracy compared to traditional machine learning models and capture local spatial features well, they still struggle to capture long-range dependencies and global context (\cite{geng2018survey}). Consequently, CNNs often perform poorly, especially for target structures that show significant inter-patient variation in size, shape, and texture (\cite{yu2023techniques}). With the introduction of Transformers and their strong performance in NLP tasks, researchers have begun applying them to vision tasks. Vision Transformers (ViTs) excel in capturing global context and understanding long-range dependencies (\cite{dosovitskiy2020image}). However, ViTs lack the local perception and inductive bias of convolutions to handle positional variations in images (\cite{lee2026survey}). 

Thus, growing interest in incorporating CNN inductive biases into ViT architectures has led to the emergence of CNN-Transformers. TransUNet (\cite{chen2021transunet}) was an early influential approach that attempted to combine a Transformer with a U-Net for medical image segmentation, capturing global features while relying on the CNN for precise localization. Subsequently, UNETR (\cite{hatamizadeh2022unetr}) advanced this idea by proposing a fully Transformer-based encoder for 3D medical image segmentation. Extending this line of work, SwinUNETR (\cite{hatamizadeh2021swin}) incorporated efficient shifted-window attention from the Swin Transformer, achieving strong performance with reduced computational complexity. However, the self-attention mechanism in Transformers is computationally expensive for high-resolution 3D medical images because its complexity scales quadratically with input sequence length. In addition, ViTs' performance may depend on the amount of training data, as they commonly require large amounts of data to learn (\cite{palanisamy2025transformers}).

In recent years, with the development of selective State Space Models (SSMs) (\cite{gu2023mamba}), Mamba has emerged as a viable alternative for capturing long-range dependencies in image segmentation. Mamba is more computationally efficient than Transformers because it delivers higher inference throughput and scales linearly with sequence length. Because of its efficiency, Mamba is ideal for the demands of 3D medical imaging, where processing large-scale structures efficiently requires both speed and precision for high-resolution volumetric data (\cite{saeed2024neuro}). Among the first attempts to apply Mamba architectures were Vision Mamba (\cite{zhu2024vision}) for general vision tasks and VM-UNet (\cite{ruan2024vm}) for medical image segmentation. Despite all these advancements in deep learning model design, there are several challenges in glioma segmentation that even state-of-the-art (SOTA)  models struggle with (\cite{khan2023survey}). 

First, the nature of glioma tumors, including severe class imbalance and uncertainty in location, size, and shape, causes difficulty in accurate segmentation. Also, boundaries are often ambiguous due to the simultaneous presence of multiple tumor subregions, such as edema, necrosis, and enhancing core (\cite{shoushtari2025current}, \cite{wu2026recent}).  

Second, most of the papers used pre-treatment glioma datasets for model training, but limited previous research has assessed segmentation of post-treatment gliomas, which is essential for disease monitoring, evaluating treatment efficacy, and planning additional treatments (\cite{dorfner2025review}, \cite{lotan2022development}). Post-treatment brain alterations, such as resection cavities (RCs), gliosis, and radiation changes, further complicate the design of a clinically useful segmentation model. Accurate delineation of resection cavities is crucial for patients receiving adjuvant radiotherapy upon surgical resection. Additionally, given the imaging similarity between necrotic cores and resection cavities, distinguishing them would be essential for evaluating total tumor volumes (\cite{de20242024}). An AI tool that could accurately segment both pre- and post-treatment glioma tumor subregions, including resection cavities, would be beneficial in daily clinical radiology practice for evaluating treatment response, clinical trials, and adjuvant radiation therapy planning, which is a laborious and time-consuming manual process. 

Finally, deep learning models are still under development for clinical use. The main cause is the “black box” nature of deep models, defined by the difficulty of explaining their internal decision process to humans. This is mainly because these models include millions of extracted feature maps and parameters in each internal layer, which are assumed to contain meaningful information about the input problem and its possible solution (\cite{shwartz2017opening}). The inherent opacity of deep learning methods makes their full comprehension highly challenging, even for domain experts, thus substantially restricting the deployment of such “black box” models in sensitive medical contexts. 

In response, explainable AI (XAI) has gained increasing research attention as a means of justifying the decision-making processes underlying DL models. Although XAI does not enhance model performance, it plays a vital role in ensuring clinical safety and fostering trust among end users, including surgeons and radiologists, by enabling deep learning models to articulate their internal reasoning in human-interpretable terms (\cite{saeed2024neuro}). However, most studies have explored XAI methods for brain tumor detection and classification rather than 3D segmentation (\cite{afnaan2025boosting}, \cite{pacal2024enhancing}).

To address these challenges, we propose a cascaded framework with a hybrid Mamba-CNN model (CMC-Seg) for segmenting pre- and post-treatment diffuse gliomas. The CMC-Seg incorporates multi-direction Mamba and convolutional blocks in the encoder, with a CNN-based decoder linked to the encoder through skip connections at various resolutions. Combining Mamba and convolutional blocks allows the model to capture both local and global features. Moreover, we use cross-attention so the model can selectively focus on relevant features across different representation spaces. Finally, we use Gradient-weighted Class Activation Mapping (Grad-CAM) to provide qualitative, class-specific saliency maps of the model’s predictions for multimodal MRI datasets.  

In summary, our work's main contributions are:

1. A cascaded multi-step framework in which a coarse segmentation model extracts tumor-centered image patches, which are then fed to CMC-Seg for fine-grained segmentation, to maximize accuracy while keeping the computational budget low. 

2. Utilization of a multi-direction 3D Mamba (MDM) block tailored for volumetric data, capturing long-range spatial dependencies along the axial, coronal, and sagital axes simultaneously.

3. Cross-attention-guided skip connection. At each decoder stage, we replace plain skip concatenation with a cross-attention mechanism that enables high-level global context from the deepest encoder stage to selectively guide feature reuse at every resolution.

4. Training and evaluation on both pre- and post-treatment glioma cases. We train and evaluate CMC-Seg against six baselines on BraTS 2025 (\cite{aboian_2025_15094823}) glioma cases to see their segmentation performance in the presence of the RC class. 

5. Explainability via gradient-based saliency. We apply Grad-CAM to the network's convolutional blocks, producing voxel-wise saliency maps, to assess whether CMC-Seg predictions rely on clinically reasonable anatomical evidence rather than spurious correlations.

\section{Related work}
\label{sec:headings}

In recent years, researchers have investigated a range of deep learning models for glioma segmentation. These efforts range from architecture optimization to adding XAI methods. The 3D U-Net extended the encoder-decoder paradigm with skip connections to volumetric data, and remains a standard baseline for dense biomedical prediction tasks (\cite{diana2025review}).

SegResNet won the BraTS 2018 challenge as a semantic segmentation network for brain tumor segmentation from multimodal 3D MRIs. In limited-data settings, the extra VAE branch helped regularize the shared encoder, boosting performance and enabling consistently high training accuracy (\cite{myronenko20183d}). In 2022, SwinUNETR was presented as a new architecture for multi-modal MRI semantic segmentation of brain tumors. This model employs a CNN-based decoder that links to the encoder via skip connections at various levels, with a Swin Transformer as the encoder in a U-shaped network architecture. This model performed competitively in the testing phase of the BraTS 2021 challenge and ranks among the top-performing techniques in the validation phase (\cite{hatamizadeh2021swin}). Xing et al. proposed SegMamba, the first general 3D medical image segmentation model based on Mamba. To improve the sequential mapping for 3D features, they developed a tri-oriented Mamba (ToM) module. They also developed a gated spatial convolution (GSC) module to model spatial relations before the ToM module. Additionally, they develop a feature-level uncertainty estimation (FUE) module to improve the skip connections of multi-scale features. SegMamba demonstrates its segmentation performance on BraTS 2023 by achieving the highest Dice score and 95\% Hausdorff distance (HD95) on whole tumor (WT), tumor core (TC), and enhancing tumor (ET) (\cite{xing2024segmambalongrangesequentialmodeling}). Lotan et al. designed an autoencoder regularization--cascaded anisotropic model for pre-operative and post-operative glioma segmentation. The autoencoder regularization--cascaded anisotropic model achieved mean Dice scores of 0.83, 0.84, and 0.72 for WT, TC/RC, and ET subregions, respectively, across both preoperative and postoperative follow-up cases (\cite{lotan2022development}). Zeineldin et al. presented a new hybrid ViT-CNN for reliable and precise glioma segmentation in brain MRI data. Their proposed approach, TransXAI, makes neural networks transparent by producing heatmaps that are understandable by surgeons. After localizing brain tumors, TransXAI uses a post-hoc explanation method that provides visual interpretation without changing the network design or affecting accuracy (\cite{zeineldin2024explainable}). Lamrani et al. introduced an explainable deep learning framework to detect, localize automatically, and segment sub-regions of brain tumors. The proposed architecture merges U-Net with dense blocks as its backbone, intended to strengthen feature reuse, enhance gradient flow, and improve overall segmentation accuracy. Visual explanations generated via Grad-CAM highlight the effectiveness of the proposed XAI approach, contributing to greater trust in and understanding of the model's behavior (\cite{lamrani2026enhanced}).

Tehsin et al. recently developed an attention-guided segmentation model to improve the accuracy and interpretability of brain tumor segmentation across the Kaggle Brain MRI, Figshare Brain Tumor, and BraTS 2020 datasets. Their pipeline integrates standardized MRI preprocessing, a hybrid Mamba encoder, a dual-branch decoder incorporating attention mechanisms, cross-scale gated fusion, and boundary-aware segmentation. This unified architecture was designed to achieve precise volumetric delineation while producing clinically interpretable results. Evaluation of the proposed model demonstrated greater overlap with anatomical regions, more clearly defined tumor boundaries, statistically significant improvements over established baselines, and superior performance and computational efficiency under domain shifts across multiple MRI datasets (\cite{tehsin2026explainable}). Despite these advances, notable gaps still exist in ensuring model robustness, interpretability, and real-world clinical applicability. Another common limitation across most related studies is the lack of post-treatment glioma datasets. Table~\ref{tab:related_works} summarizes related works and their advantages and disadvantages.

\begin{table*}[!t]
	\centering
	\caption{Summary of some related works.}
	\label{tab:related_works}
	\small
	\setlength{\tabcolsep}{4pt}
	\renewcommand{\arraystretch}{1.2}
	\begin{tabularx}{\textwidth}{@{}L{2.15cm} L{2.75cm} L{2.9cm} Y Y@{}}
		\toprule[1.2pt]
		\rowcolor{gray!20}
		\textbf{Study} & \textbf{Model backbone} & \textbf{Dataset} & \textbf{Pros} & \textbf{Cons} \\
		\midrule[0.8pt]
		(\cite{myronenko20183d}) &
		CNNs (SegResNet) &
		BraTS 2018 &
		Additional VAE branch\newline
		Improved generalization under limited data &
		No explicit long-range dependency modeling\newline
		Absence of post-treatment data\newline
		Absence of XAI \\
		\addlinespace[2pt]
		
		\rowcolor{gray!8}
		(\cite{ma2024dtasunet}) &
		Hybrid Transformer-CNN (DTASUnet) &
		BraTS 2018,\newline
		BraTS 2020 &
		Capturing both local features and global context\newline
		Interpretability &
		Computational cost\newline
		Absence of post-treatment data \\
		\addlinespace[2pt]
		
		(\cite{hatamizadeh2021swin}) &
		Hybrid Transformer-CNN (Swin UNETR) &
		BraTS 2021 &
		Hierarchical shifted-window attention\newline
		Capturing both local features and global context &
		Data-hungry without pretraining\newline
		Absence of post-treatment data\newline
		Absence of XAI \\
		\addlinespace[2pt]
		
		\rowcolor{gray!8}
		(\cite{liu2023attention}) &
		CNN (AMMGS) &
		BraTS2020,\newline
		BraTS2019 &
		Focus on small-intensity dissimilarity &
		Absence of post-treatment data\newline
		Absence of XAI \\
		\addlinespace[2pt]
		
		(\cite{xing2024segmambalongrangesequentialmodeling}) &
		Hybrid Mamba-CNN (SegMamba) &
		BraTS 2023 &
		Capturing both local features and global context\newline
		Global receptive field at linear complexity &
		Absence of post-treatment data\newline
		Absence of XAI \\
		\addlinespace[2pt]
		
		\rowcolor{gray!8}
		(\cite{jin2026dm}) &
		Hybrid Mamba-CNN (DM-UNet) &
		BraTS 2023 &
		Low computational cost\newline
		Capturing both local features and global context\newline
		Global receptive field at linear complexity &
		Limitations in handling small lesions, low-contrast regions, and intricate boundary structures\newline
		Absence of post-treatment data\newline
		Absence of XAI \\
		\addlinespace[2pt]
		
		(\cite{zeineldin2024explainable}) &
		Hybrid Transformer-CNN (TransXAI) &
		BraTS 2019 &
		Capturing both local features and global context\newline
		Interpretability &
		Evaluated only on 2D data\newline
		Absence of post-treatment data \\
		\addlinespace[2pt]
		
		\rowcolor{gray!8}
		(\cite{lotan2022development}) &
		CNN (AR-CA) &
		BraTS 2019, private data &
		Evaluation on both pre- and post-treatment data\newline
		Autoencoder regularization &
		No explicit long-range dependency modeling\newline
		Absence of XAI \\
		\addlinespace[2pt]
		
		(\cite{lamrani2026enhanced}) &
		CNN (Dense-U-Net) &
		BraTS 2020,\newline
		Kaggle Brain MRI &
		Adaptive loss weighting\newline
		Spatially focused preprocessing\newline
		Use of Dense blocks\newline
		Interpretability &
		No explicit long-range dependency modeling\newline
		Absence of post-treatment data \\
		\addlinespace[2pt]
		
		\rowcolor{gray!8}
		(\cite{tehsin2026explainable}) &
		CNN--Transformer--Mamba &
		Kaggle Brain MRI, Figshare Brain Tumor, and BraTS2020 &
		Capture local texture, long-range context, and spatial boundary cues\newline
		Multi-dataset validation\newline
		Interpretability &
		Evaluated only on 2D data\newline
		Absence of post-treatment data \\
		\bottomrule[1.2pt]
	\end{tabularx}
	
	\vspace{3pt}
	\parbox{\textwidth}{\footnotesize
		\textit{Note.} XAI, explainable AI; CNN, convolutional neural network; VAE, variational autoencoder; MRI, magnetic resonance imaging.}
\end{table*}

\section{Materials and methods}

In this section, we present the technical framework of our proposed method for diffuse glioma segmentation. We first describe the dataset characteristics, splitting, and preprocessing steps. Second, we define the overall architecture and present the mathematical formulation of our two main components: the MDM block and cross-attention. Finally, we explain the training process, evaluation metrics, and statistical analysis. Fig. 1 shows the overall workflow.

\begin{figure*}[!t]
	\centering
	\includegraphics[width=\textwidth,height=0.72\textheight,keepaspectratio]{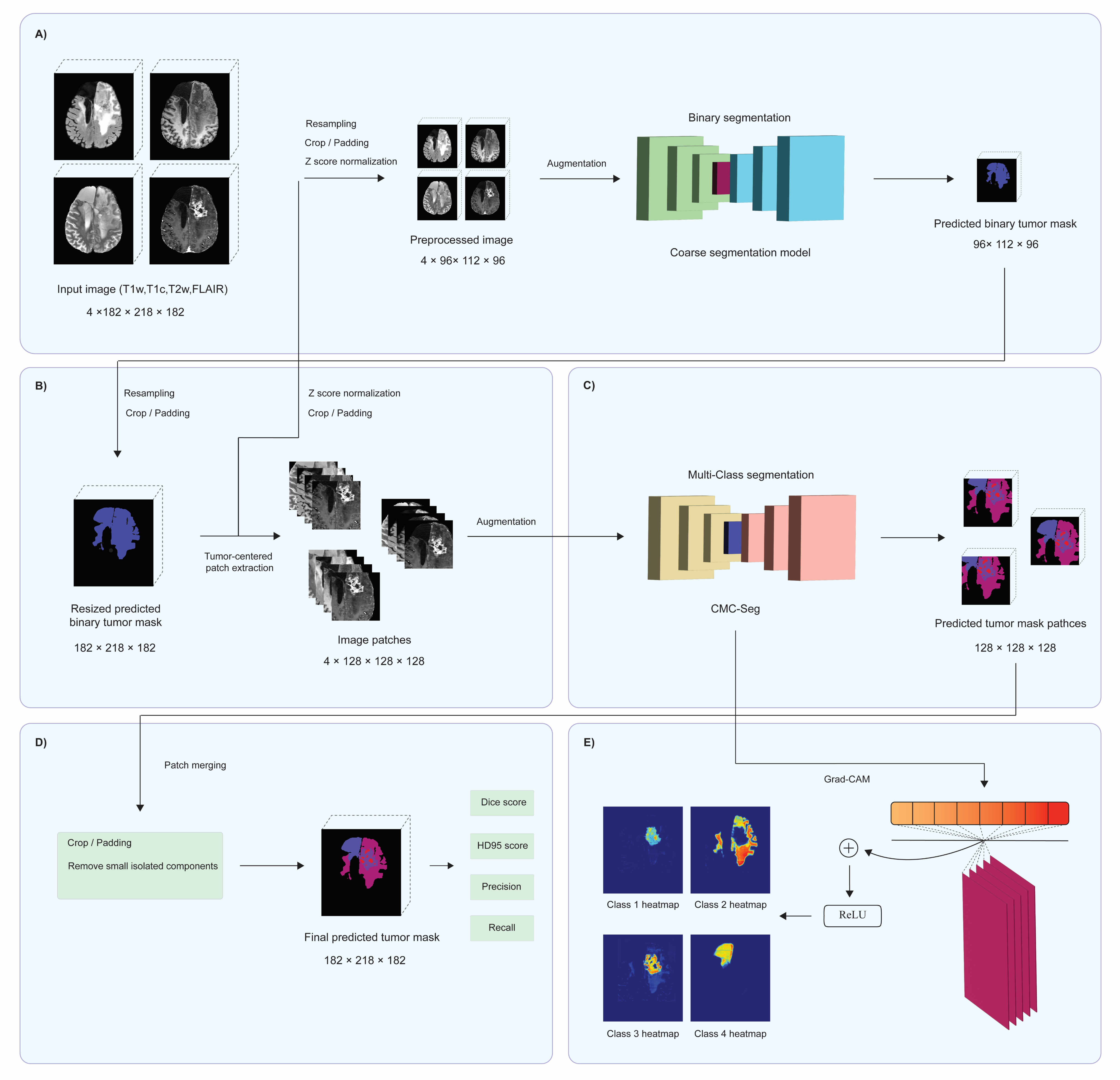}
	\caption{The overall workflow of proposed framework. A) Training and utilization of a coarse segmentation model to approximately detect the ROI. B) Patch extraction from the original input image according to the ROI. C) Training and utilization of CMC-Seg to predict mask patches. D) Merging the predicted mask patches to reconstruct the full-size masks and evaluate the segmentation performance. E) Conducting explainability experiments.}
\label{FIG:1}
\end{figure*}

\subsection{Dataset}
We used BraTS 2025 Task one data that includes 2872 patients with diffuse glioma, which was acquired over multiple institutions and includes routine pre- and post-treatment multi-parametric MRI (mpMRI) (\cite{aboian_2025_15094823}). Since gliomas are heterogeneous, no single sequence captures all tumor components; therefore, four MRI modalities, including skull-stripped native T1w, T1c, T2w, and FLAIR, were used as input images. The original isotropic voxel spacing was 1 mm, and the image size was 182×218×182. Tumor masks included four classes: non-enhancing tumor core (NETC), surrounding non-enhancing FLAIR hyperintensity (SNFH), ET, and RC. Because of clinical importance, two regions, including TC and WT, are also considered for evaluation. The TC consists of NETC and ET, representing the lesion that is usually removed during surgical resection. The WT consists of NETC, SNFH, and ET, representing the whole extent of the tumor.

\subsection{Preprocessing}

Before preprocessing, we randomly partitioned the dataset at the subject level into a development set (90\%) and a held-out test set (10\%), stratified by treatment status to preserve the original cohort distribution in the test set. The development set was then split into training (80\% of the total) and validation (10\% of the total) subsets. 

\begin{table*}[!t]
	\centering
	\caption{Prevalence and mean volume $\pm$ standard deviation of each tumor class.}
	\label{tab:tumor_stats}
	\small
	\setlength{\tabcolsep}{6pt}
	\renewcommand{\arraystretch}{1.15}
	\begin{tabular}{@{}l l c c c@{}}
		\toprule[1.2pt]
		\rowcolor{gray!25}
		\textbf{Tumor class} & \textbf{Subgroup} & \textbf{Development dataset} & \textbf{Test dataset} & \textbf{Total dataset} \\
		\midrule[0.8pt]
		\rowcolor{gray!15}
		\multicolumn{5}{@{}l}{\textbf{Prevalence (\%)}} \\
		NETC & Pre-treatment  & 96.54 & 96.75 & 96.56 \\
		& Post-treatment & 43.45 & 44.51 & 43.55 \\
		& Total          & 66.62 & 66.9  & 66.64 \\
		\addlinespace[3pt]
		\rowcolor{gray!8}
		SNFH & Pre-treatment  & 99.91 & 100   & 99.92 \\
		\rowcolor{gray!8}
		& Post-treatment & 99.79 & 100   & 99.81 \\
		\rowcolor{gray!8}
		& Total          & 99.85 & 100   & 99.86 \\
		\addlinespace[3pt]
		ET   & Pre-treatment  & 97.43 & 96.75 & 97.36 \\
		& Post-treatment & 74.95 & 79.88 & 75.45 \\
		& Total          & 84.76 & 87.11 & 84.99 \\
		\addlinespace[3pt]
		\rowcolor{gray!8}
		RC   & Pre-treatment  & -     & -     & -     \\
		\rowcolor{gray!8}
		& Post-treatment & 84.76 & 84.76 & 84.76 \\
		\rowcolor{gray!8}
		& Total          & 47.78 & 48.43 & 47.84 \\
		\midrule[0.8pt]
		\rowcolor{gray!15}
		\multicolumn{5}{@{}l}{\textbf{Mean volume (cm$^3$)}} \\
		NETC & Pre-treatment  & 15.03\,$\pm$\,21.2   & 12.85\,$\pm$\,17.01 & 14.81\,$\pm$\,20.83 \\
		& Post-treatment & 3.86\,$\pm$\,6.69    & 5.28\,$\pm$\,7.89   & 4\,$\pm$\,6.84 \\
		& Total          & 10.92\,$\pm$\,18.16  & 9.97\,$\pm$\,14.69  & 10.83\,$\pm$\,17.84 \\
		\addlinespace[3pt]
		\rowcolor{gray!8}
		SNFH & Pre-treatment  & 60.53\,$\pm$\,42.41  & 57.82\,$\pm$\,40.81 & 60.26\,$\pm$\,42.25 \\
		\rowcolor{gray!8}
		& Post-treatment & 49.58\,$\pm$\,43.74  & 51.79\,$\pm$\,41.73 & 49.8\,$\pm$\,43.53 \\
		\rowcolor{gray!8}
		& Total          & 54.36\,$\pm$\,43.49  & 54.37\,$\pm$\,41.37 & 54.36\,$\pm$\,43.28 \\
		\addlinespace[3pt]
		ET   & Pre-treatment  & 22.32\,$\pm$\,18.09  & 19.31\,$\pm$\,16.34 & 22.03\,$\pm$\,17.95 \\
		& Post-treatment & 11.78\,$\pm$\,18.16  & 11.99\,$\pm$\,17.27 & 11.8\,$\pm$\,16.82 \\
		& Total          & 17.07\,$\pm$\,18.22  & 15.48\,$\pm$\,17.2  & 16.9\,$\pm$\,18.12 \\
		\addlinespace[3pt]
		\rowcolor{gray!8}
		RC   & Pre-treatment  & -                    & -                   & - \\
		\rowcolor{gray!8}
		& Post-treatment & 15.04\,$\pm$\,21.14  & 16.84\,$\pm$\,22.34 & 15.23\,$\pm$\,21.26 \\
		\rowcolor{gray!8}
		& Total          & 15.04\,$\pm$\,21.14  & 16.84\,$\pm$\,22.34 & 15.23\,$\pm$\,21.26 \\
		\bottomrule[1.2pt]
	\end{tabular}
	
	\vspace{3pt}
	\parbox{\textwidth}{\footnotesize
		NETC, non-enhancing tumor core; SNFH, surrounding non-enhancing FLAIR hyperintensity; ET, enhancing tumor; RC, resection cavity.}
\end{table*}

The volumetric data and distribution of each tumor class are provided in Table~2. Before training, the images underwent a series of preprocessing steps. First, data integrity was assessed, and patients were excluded if they exhibited corrupted or missing image or mask data. All 2,872 patients passed this quality-control assessment; thus, no patients were excluded on the basis of data integrity. All input images (T1w, T2w, FLAIR, T1c) were normalized using $z$-score standardization to improve training stability and convergence. In $z$-score standardization, we used only statistics from each volume; no information was shared between the training, validation, and test sets.

\begin{equation}
	Z = \frac{x-\mu}{\sigma}
\end{equation}

where $Z$ represents the $z$-score, $x$ represents input data, $\mu$ represents the mean, and $\sigma$ represents the standard deviation. To balance computational efficiency with preservation of relevant anatomical context, a coarse segmentation model was first employed to localize the tumor region. Our coarse segmentation model is a CNN architecture that receives input images (T1w, T2w, FLAIR, T1c), which were resized to $96 \times 112 \times 96$ with linear interpolation. The original multiclass segmentation masks were converted into binary masks by assigning all non-background voxels to the foreground class. The coarse CNN was trained using the training subset and the corresponding binary segmentation targets. The predicted binary masks were resized to $182 \times 218 \times 182$ using nearest-neighbor interpolation and used for region-of-interest (ROI) detection. After detection of the ROI, a 15~mm safe margin was added in each direction of each axis then image patches with a size of $128 \times 128 \times 128$ from the final ROI were extracted. This approach reduces computational requirements while retaining sufficient anatomical context for subsequent segmentation. Finally, to improve model generalization, address limited data diversity, and class imbalance, on-the-fly 3D data augmentation during training was utilized. Augmentations included random flipping along one of the three spatial axes, gamma transformations, additive Gaussian noise, and intensity scaling and shifting.

\subsection{Network Architecture}

\begin{algorithm}[!t]
	\caption{CMC-Seg model forward pass.}
	\label{alg:cmcseg_forward}
	\footnotesize
	\begin{algorithmic}[1]
		\Require MRI volumes including T1w, T1c, T2w and FLAIR,
		$x \in \mathbb{R}^{C\times D\times H\times W}$
		\Ensure  Tumor mask, $y \in \mathbb{R}^{K\times D\times H\times W}$
		
		\State $s_0 \gets \text{ResidualBlock}(x)$, $F$ channels
		\State $f_1 \gets \text{PatchEmbedding}(x)$, $2F$ channels
		\State $f_1 \gets \text{MDMBlock}(f_1)$
		\State $s_1 \gets \text{ResidualBlock}(f_1)$
		
		\For{$i = 2$ \textbf{to} $5$}
		\State $f_i \gets \text{Conv3D}_{3\times3\times3,\ \text{stride }2}(s_{i-1})$,
		$2F\times 2^{i-1}$ channels
		\State $f_i \gets \text{MDMBlock}(f_i)$
		\State $s_i \gets \text{ResidualBlock}(f_i)$
		\EndFor
		
		\State $d_0 \gets s_4$
		
		\For{$i = 1$ \textbf{to} $4$}
		\State $d_i \gets \text{ConvTranspose3D}_{4\times4\times4,\ \text{stride }2}(d_{i-1})$,
		$2F\times 2^{3-i}$ channels
		\State $Q_i \gets \text{PatchEmbedding}(s_{4-i})$,
		$2F\times 2^{3-i}$ channels
		\State $KV_i \gets \text{PatchEmbedding}(d_0)$,
		$2F\times 2^{3-i}$ channels
		\State $h_i \gets \text{CrossAttention}(Q_i, KV_i)$
		\State $h_i \gets \text{Reshape}(h_i)$
		\State $s_{4-i} \gets \text{GroupNorm}(s_{4-i} + h_i)$
		\State $d_i \gets \text{Concatenate}(d_i,\, s_{4-i})$,
		$4F\times 2^{3-i}$ channels
		\State $d_i \gets \text{ResidualBlock}(d_i)$,
		$2F\times 2^{3-i}$ channels
		\EndFor
		
		\State $d_4 \gets \text{Conv3D}_{3\times3\times3,\ \text{stride }1}(d_4)$, $F$ channels
		\State $d_4 \gets \text{GELU}(d_4)$
		\State $d_4 \gets \text{GroupNorm}(d_4)$
		\State $y \gets \text{Conv3D}_{1\times1\times1,\ \text{stride }1}(d_4)$, $K$ channels
		
		\State \Return $y$
	\end{algorithmic}
	\vspace{2pt}
	\parbox{\linewidth}{\footnotesize
		\textit{Forward pass of the CMC-Seg model. The input image is MRI volumes including T1w, T1c, T2w, and FLAIR ($C=4$). The encoder combines residual blocks with MDM blocks across four resolution levels. The decoder fuses upsampled features with cross-attention-refined encoder skip features. A final convolutional head outputs a $K$-class tumor mask.}}
\end{algorithm}

\begin{figure*}
	\centering
	\includegraphics[width=\textwidth]{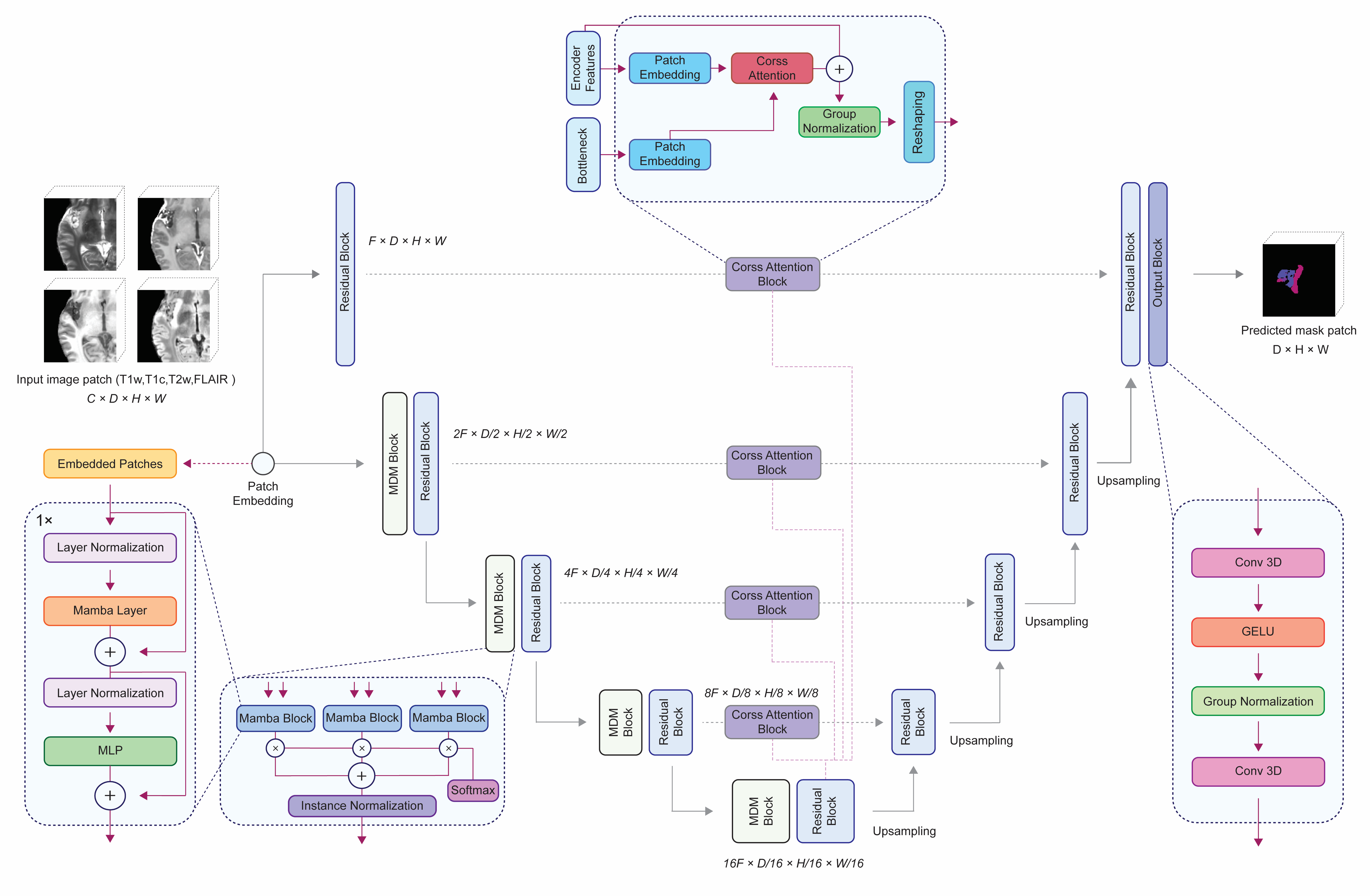}
	\caption{Detailed architecture of CMC-Seg. H, W, D correspond to the spatial dimensions; C is the channel size of the input image, and F is the base feature size. An input 3D MRI volume is processed through an initial patch embedding and four encoder stages. Each stage consists of an MDM and a residual block, followed by a convolution layer with a stride of 2. The MDM block scans the flattened image along three different axes plus their reverse directions. The decoder symmetrically upsamples the features, fuses them with filtered encoder features via cross-attention, and then the fused features go through the residual block.}
	\label{FIG:2}
\end{figure*}

CMC-Seg is a U-Net--style encoder--decoder for volumetric semantic segmentation that utilizes a hybrid Mamba-CNN design. After downsampling the image at each level of the encoder, the image is fed to an MDM block followed by a residual convolutional block. The decoder part is a pure CNN, which consists of an upsampling layer followed by a concatenation layer and residual blocks, respectively. The encoder features are propagated to the decoder through a set of cross-attention skip-fusion modules that inject global context from the deepest encoder stage into each decoder-side skip feature before concatenation (Algorithm~1). Fig.~2 shows a detailed view of the CMC-Seg architecture.

\subsubsection{Patch Embedding and Positional Encoding}

The input volume has the shape of $x \in \mathbb{R}^{B \times C \times D \times H \times W}$ where $B$ and $C$ are the batch size and channel size, respectively, as well as $D$, $H$, $W$ correspond to the spatial dimensions. The input is first projected into a feature grid using a 3D convolutional patch-embedding layer with kernel size and stride of 2. Rather than a single flattened positional embedding, axis-factorized positional encodings were used, including three independent learnable embedding tables $E_D \in \mathbb{R}^{D_p \times C_p}$, $E_H \in \mathbb{R}^{H_p \times C_p}$, and $E_W \in \mathbb{R}^{W_p \times C_p}$ (one per spatial axis, truncated-normal initialized) are broadcast-summed onto the projected feature grid:

\begin{equation}
	y = \mathrm{Conv3D}(x) + E_D + E_H + E_W
\end{equation}

This factorization reduces the parameter count relative to a full 3D positional embedding table form while still providing axis-resolved positional information.

\subsubsection{MDM Block}

The core sequence-mixing unit at every encoder stage is the MDM block, which extends a standard SSM layer (Mamba) with a multi-direction scan to address the lack of a canonical 1D traversal order in volumetric data. Given a feature map $x \in \mathbb{R}^{B \times C \times D \times H \times W}$, the module flattens the spatial grid into token sequences along three orderings, including depth-major ($D \times H \times W$), height-major ($H \times D \times W$), and width-major ($W \times D \times H$) --- obtained by axis permutation before flattening. 

Each ordering exposes a different spatial adjacency structure to the Mamba scan, allowing the model to capture dependencies along all three principal axes rather than only the one implied by a fixed order. For each of the three orderings, the sequence and its reverse are passed through a shared Mamba block (pre-norm Mamba layer + pre-norm Multi-Layer Perceptron (MLP) with residual connections) where $x$, $\hat{x}$, and $z$ denote the input, intermediate, and output feature representations, respectively; $\mathrm{LN}$ and Mamba Layer denote layer normalization and the Mamba sequence-modeling layer, respectively. $W$ and $b$ represent learnable weights and biases:

\begin{equation}
	\hat{x} = x + \mathrm{Mamba\ Layer}(\mathrm{LN}_1(x))
\end{equation}
\begin{equation}
	z = \hat{x} + \mathrm{MLP}(\mathrm{LN}_2(\hat{x}))
\end{equation}
\begin{equation}
	z = \mathrm{Mamba\ block}(x)
\end{equation}
\begin{equation}
	\mathrm{MLP}(u) = W_2 \cdot \mathrm{GELU}(W_1 u + b_1) + b_2
\end{equation}

The reverse branch's output is flipped back to the original token order and added to the forward branch, yielding a direction-invariant representation for that axis where $s$ denotes the input feature map, $x^{(a)}$ represents the sequence obtained using the $a$ spatial scanning order, and $a \in \{1,2,3\}$ indexes the three scanning directions:

\begin{equation}
	x^{(1)} = \mathrm{flatten}_{DHW}(s), \quad x^{(2)} = \mathrm{flatten}_{HDW}(s), \quad x^{(3)} = \mathrm{flatten}_{WDH}(s)
\end{equation}
\begin{equation}
	\left[ z_{\mathrm{fwd}}^{(a)}; z_{\mathrm{rev}}^{(a)} \right] = \mathrm{Mamba\ block}\left( \left[ x^{(a)}; \mathrm{flip}(x^{(a)}) \right] \right)
\end{equation}
\begin{equation}
	z^{(a)} = z_{\mathrm{fwd}}^{(a)} + \mathrm{flip}(z_{\mathrm{rev}}^{(a)}), \quad a \in \{1,2,3\}
\end{equation}

This is repeated independently for all three axis orderings (six total scan directions: 3 axes $\times$ 2 directions), and each of the three resulting feature maps is reshaped back to $B \times F \times D \times H \times W$, where $F=32$ is the base feature size. The three axis-wise outputs are then combined via a learned, softmax-normalized, per-channel convex combination:

\begin{equation}
	p^a = \mathrm{reshape}_{B \times F \times D \times H \times W}(z^a)
\end{equation}
\begin{equation}
	\tilde{w} = \mathrm{softmax}(w), \quad w \in \mathbb{R}^{3 \times F}
\end{equation}
\begin{equation}
	\mathrm{MDM\ block}(s) = \mathrm{InstanceNorm3D}\left( \sum_{k=1}^3 \tilde{w}_k \odot p_k \right)
\end{equation}

where $w$ is a learnable parameter initialized to zero. An instance normalization layer is applied to the fused output to stabilize training. This design allows the network to weight each spatial scan direction adaptively and per-channel, rather than assuming a fixed traversal order is optimal for all feature channels.

\subsubsection*{3.3.3 Encoder}
The encoder consists of four stages with increasing channel width ($2F$, $4F$, $8F$, $16F$) and decreasing spatial resolution. At each stage:
\begin{enumerate}
	\item MDM block (with $d$ state scaled per stage: 16, 32, 64, 128) models the global context.
	\item A residual convolutional block (Residual block: GroupNorm $\rightarrow$ GELU $\rightarrow$ Conv3D $\rightarrow$ GroupNorm $\rightarrow$ GELU $\rightarrow$ Conv3D, with an identity or $1\times1\times1$-convolution residual path) refines local features.
	\item A $3\times3\times3$ convolution layer with a stride of $2\times2\times2$ that downsamples to the next resolution while doubling the channel count.
\end{enumerate}
A parallel shallow residual stem operates on the input at full resolution to preserve a fine-grained skip connection ($F$ channels) that bypasses the patch-embedding downsampling entirely. The output of the fourth encoder stage (resolution $D/16 \times H/16 \times W/16$, $16F$ channels) serves as the shared global-context source for all cross-attention skip-fusion operations in the decoder.

\subsubsection*{3.3.4 Cross-Attention Skip Fusion}
Conventional U-Net skip connections copy encoder features directly to the decoder at matching resolutions, limiting their receptive field to that stage's local context. To address this, each of the four skip levels is additionally refined by a cross-attention operation (MONAI's CrossAttentionBlock) (\cite{cardoso2022monai}) before concatenation with the corresponding decoder feature map:
\begin{itemize}
	\item Queries are formed by patch embedding, where $s_{4-i}$ denotes the skip feature at level $4-i$.
	\item Keys/values are formed by patch embedding, where $b = s_4$ denotes the bottleneck feature map, each is projected to a channel dimension matching the corresponding query.
\end{itemize}
\begin{equation}
	\hat{s}_{4-i} = \text{GroupNorm}\left(s_{4-i} + \text{reshape}\left(\text{CrossAttn}\left(q=\text{Embed}(s_{4-i}), kv=\text{Embed}(b)\right)\right)\right), \quad i \in \{1,2,3,4\} \tag{13}
\end{equation}

This injects deep, semantically rich bottleneck context directly into shallow/mid-level skip features, letting the decoder access long-range information at every resolution rather than only through the sequential upsampling path. We use four such cross-attention modules (4, 4, 2, and 2 attention heads, respectively, with decreasing channel widths of $8F$, $4F$, $2F$, and $F$), each followed by a residual add and GroupNorm before concatenation with the transposed-convolution decoder output and a residual block.

\subsubsection*{3.3.5 Decoder}
The decoder mirrors the encoder with four stages of $4\times4\times4$ transposed convolutions with a stride of $2\times2\times2$ for upsampling, each followed by concatenation with the corresponding cross-attention--refined skip feature and a residual convolutional block. The final decoder stage produces an $F$ channels feature map, refined by an additional Conv3D--GELU--GroupNorm block, before a $1\times1\times1$ convolutional output head produces the final per-voxel class logits.

\subsection*{3.4 Training}
The models were trained on Kaggle (Python 3.12) with $2\times$ NVIDIA T4 GPUs (15 GB each), 30 GB of system RAM (PyTorch 2.10), using a batch size of 2 to balance GPU memory constraints and stable gradient updates. The segmentation models were trained for 100 epochs using the AdamW optimizer with an initial learning rate of $5\times10^{-5}$, $\beta_1=0.9$, $\beta_2=0.95$, and a weight decay of 0.05. To improve optimization stability, the learning rate was linearly warmed up during the initial 8\% of the total training steps, followed by cosine annealing to a minimum learning rate of $1\times10^{-6}$. Mixed-precision training with automatic loss scaling was employed to reduce GPU memory consumption and improve computational efficiency, while gradient norms were clipped to 1.0 to improve training stability. 
We employed a combined loss function, integrating Tversky loss (\cite{salehi2017tversky}) and categorical focal cross-entropy (CFCE) loss (\cite{lin2017focal}), where $C$ denotes the total number of segmentation classes and equal weight is given to each loss ($\omega_1 = \omega_2 = 1.0$). In Tversky loss, $\text{TP}_c$, $\text{FP}_c$, and $\text{FN}_c$ denote the numbers of true-positive, false-positive, and false-negative voxels for class $c$, respectively. The coefficients $\alpha$ and $\beta$ control the relative penalties assigned to false-positive and false-negative predictions, respectively. The Tversky loss was configured with $\alpha = \neg 0.3$ and $\beta = 0.7$, assigning a higher penalty to false negatives than false positives. This configuration is useful for glioma segmentation because it emphasizes detecting tumor voxels and helps reduce the risk of under-segmentation of small or difficult tumor regions. In CFCE loss, $N$ denotes the total number of voxels, $g_{ic}$ is the one-hot encoded ground-truth label for voxel $i$ and class $c$, $p_{ic}$ is the predicted probability of class $c$ for voxel $i$, $\alpha$ is the weighting factor, and $\gamma$ is the focusing parameter. The categorical focal cross-entropy loss was configured with $\gamma = 2.0$ and $\alpha = 0.25$ to place greater emphasis on hard-to-classify voxels and reduce the influence of easily classified examples. 

\begin{equation}
	L_{\text{total}} = \omega_1 L_{\text{tversky}} + \omega_2 L_{\text{CFCE}} \tag{14}
\end{equation}
\begin{equation}
	L_{\text{tversky}} = 1 - \frac{1}{C} \sum_{c=1}^{C} \frac{\text{TP}_c}{\text{TP}_c + 0.3\text{FP}_c + 0.7\text{FN}_c} \tag{15}
\end{equation}
\begin{equation}
	L_{\text{CFCE}} = -\frac{1}{N} \sum_{i}^{N} \sum_{c=1}^{C} 0.25 g_{ic} (1-p_{ic})^2 \log(p_{ic}) \tag{16}
\end{equation}

\subsection*{3.5 Evaluation metrics}
To comprehensively evaluate the model's performance, we used metrics including Dice score, HD95, precision, and recall. We computed each metric for WT, TC, ET, NETC, SNFH, RC, and the mean across individual tumor classes. During training, we monitored and compared model performance using Intersection-over-Union (IoU) as the primary evaluation metric. In metrics $P_c$ and $G_c$ correspond to predicted and ground truth masks for class $c$, respectively. The $\text{TP}_c$, $\text{FP}_c$, and $\text{FN}_c$ denote the numbers of true-positive, false-positive, and false-negative voxels for class $c$, respectively. 
Dice score: measures the volumetric overlap between predicted and ground-truth segmentation masks.

\begin{equation}
	\text{Dice score}_c = \frac{2|P_c \cap G_c|}{|P_c| + |G_c|} = \frac{2\text{TP}_c}{2\text{TP}_c + \text{FN}_c + \text{FP}_c} \tag{17}
\end{equation}

IoU: measures the intersection relative to the union of the predicted and ground-truth segmentation regions.

\begin{equation}
	\text{IoU}_c = \frac{|P_c \cap G_c|}{|P_c \cup G_c|} = \frac{\text{TP}_c}{\text{TP}_c + \text{FN}_c + \text{FP}_c} \tag{18}
\end{equation}

HD95: quantifies boundary accuracy by measuring the distance between the surfaces of the predicted and ground-truth masks, using the 95th percentile rather than the maximum to reduce sensitivity to single-voxel outliers. The $A$ and $B$ denote the predicted and ground-truth segmentation surfaces, respectively; $d(a,B)$ represents the minimum distance between point $a \in A$ in $A$ and surface $B$, and $d(b,A)$ is defined analogously.

\begin{equation}
	\text{HD}_{95}(A,B) = \max\left\{ P_{95}\{d(a,B): a \in A\}, P_{95}\{d(b,A): b \in B\} \right\} \tag{19}
\end{equation}

Precision: measures the proportion of predicted foreground voxels that are true positives.

\begin{equation}
	\text{Precision}_c = \frac{|P_c \cap G_c|}{|P_c|} = \frac{\text{TP}_c}{\text{TP}_c + \text{FP}_c} \tag{20}
\end{equation}

Recall: measures the proportion of ground-truth foreground voxels that were correctly identified.

\begin{equation}
	\text{Recall}_c = \frac{|P_c \cap G_c|}{|G_c|} = \frac{\text{TP}_c}{\text{TP}_c + \text{FN}_c} \tag{21}
\end{equation}

When a class was absent in both the ground truth and the prediction for a patient ($|P|=0$ and $|G|=0$), Dice, precision, and recall were assigned 100\%, and HD95 was assigned 0, reflecting perfect agreement on a true negative. When a class was present in only one of the two ($|P|=0$ or $|G|=0$), Dice, precision, and recall were assigned 0, and HD95 was assigned a fixed penalty of 337.3 mm, corresponding to the maximum possible distance within the evaluated volume. We retained these cases in the metric aggregation rather than excluding them or treating them as undefined. Since tumor class RC does not occur in pre-treatment cases, we computed RC metrics only for post-treatment cases.

\subsection{Experiment and Analysis}
We benchmarked CMC-Seg against six segmentation models spanning convolutional, Transformer, and SSM paradigms, including 3D U-Net\cite{cciccek20163d}, SegResNet\cite{myronenko20183d}, SwinUNETR\cite{hatamizadeh2021swin}, UNETR\cite{hatamizadeh2022unetr}, nnFormer\cite{zhou2023nnformer}, and SegMamba\cite{xing2024segmambalongrangesequentialmodeling}. All baselines underwent the same preprocessing, data splits, training, and postprocessing protocols to isolate architectural contribution from confounding pipeline differences. Model outputs were converted to discrete label maps via voxel-wise argmax over predicted class probabilities. Then we merged the predicted mask patches of each patient to reconstruct the original mask size of $182\times218\times182$. To suppress spurious small predictions arising from noise or partial-volume effects, we applied per-class connected-component filtering: for each foreground class, 3D connected components were identified using 6-connectivity (\texttt{scipy.ndimage.label}, default structuring element), and components below a class-specific minimum voxel threshold were discarded. Minimum component sizes were set empirically per class (30 voxels for classes 1 and 3, 20 voxels for class 2, 80 voxels for class 4), reflecting differences in typical lesion sub-region size.

For each architecture, we calculate the mean $\pm$ standard deviation of per-class Dice score, HD95, precision, and recall for NETC, SNFH, ET, and RC, together with WT and TC composite scores. Statistical significance between our model and each baseline was assessed using paired Wilcoxon signed-rank tests at the patient level, with Benjamini-Hochberg false discovery rate correction applied across the resulting family of comparisons. To isolate the contribution of key architectural components, we conducted a component-wise ablation study in which each component was independently removed or modified while holding all other architectural elements, hyperparameters, and training schedule fixed. Three ablations were performed: (i) removal of the cross-attention skip fusion block, replaced with a standard skip connection, to assess the contribution of attention-based feature fusion between encoder and decoder; (ii) removal of the multi-directional Mamba scanning scheme and utilizing a one-directional Mamba block instead to assess the contribution of multi-directional scanning; and (iii) removal of both the cross-attention skip fusion block and multi-directional Mamba to assess the synergistic effect of employing them together. Each ablation variant was trained from the same initialization and data ordering as the full model, and evaluated using the identical metric suite and statistical protocol. Apart from segmentation accuracy, computational efficiency is crucial for deploying deep learning models in clinical settings. Here, we present experimental analyses of CMC-Seg complexity, comparing it with CMC-Seg variants and representative benchmark models from different architecture families. In the computational complexity analysis, we assess the parameters, Giga floating-point operations (GFLOPs), peak VRAM usage, and average inference time.

\subsection{Interpretability}
To assess whether the model's segmentation decisions are based on clinically plausible image regions, we applied Grad-CAM, adapted for 3D volumetric test data. For a given input patch and target class, we computed the gradient of the target-class summed pre-softmax logit map with respect to the activations of a chosen target convolutional layer. We obtained channel-wise importance weights by global average pooling these gradients over the spatial dimensions, and produced a class-discriminative localization map as a ReLU-thresholded weighted combination of the layer's activation maps. If the resulting map dimensions were lower than the input patch resolution, we upsampled it (trilinear interpolation) to $128 \times 128 \times 128$ and min-max normalized it to $[0, 1]$. For the explainability experiment, we selected several layers, including the bottleneck layer and the decoder's third and final residual block. We produced per-patch Grad-CAM maps independently and reassembled them into a full-volume heatmap ($182\times218\times182$) using the same spatial metadata used for patch reconstruction during inference.

\begin{equation}
	\alpha_k^c = \frac{1}{Z} \sum_i \sum_j \sum_l \frac{\partial Y^c}{\partial A_{ijl}^k}
\end{equation}

\begin{equation}
	L_{\text{Grad-CAM}}^c = \text{ReLU}\left(\sum_k \alpha_k^c A^k \right)
\end{equation}

\begin{equation}
	\text{Mean heatmap Intensity} = \frac{1}{N} \sum_{i=1}^{N} L_{\text{Grad-CAM}}^c(i)
\end{equation}

\section{Results}
\subsection{Comparison with state-of-the-art models}
To evaluate CMC-Seg comprehensively, we benchmarked it against an array of SOTA models. The competing models were categorized into three groups: CNN-based, Transformer-based, and Mamba-based. All baselines (CNN-, Transformer-, and Mamba-based) were trained and evaluated under the same conditions to ensure a thorough and fair comparison.

\begin{figure*}
	\centering
	\includegraphics[width=\textwidth]{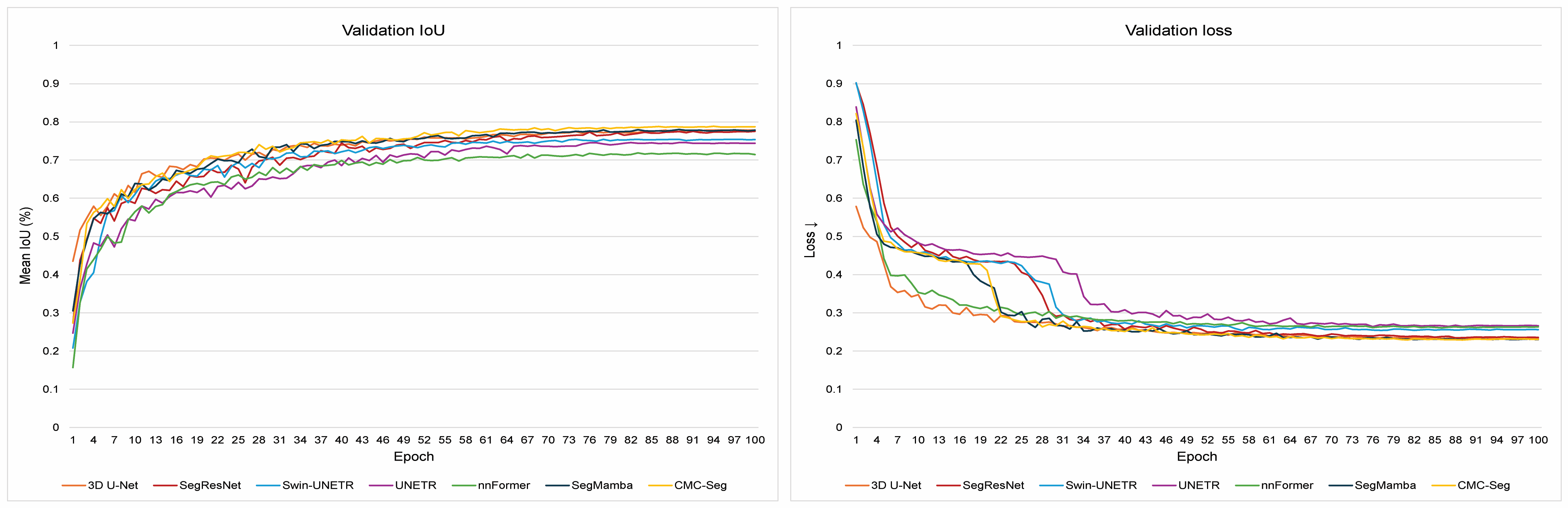}
	\caption{Validation IoU and validation loss plot of CMC-Seg and benchmark models. CMC-Seg achieves the best validation IoU and loss compared to benchmark models.}
	\label{FIG:3}
\end{figure*}

Fig.~3 shows the validation IoU and validation loss for each of the seven models over 100 training epochs. Compared to SegMamba (peak 77.9\%, final 77.8\%), 3D U-Net (peak 77.9\%, final 77.8\%), SegResNet (peak 77.6\%, final 77.5\%), Swin-UNETR (peak 75.4\%, final 75.4\%), UNETR (peak 74.6\%, final 74.4\%), and nnFormer (peak 71.9\%, final 71.5\%), CMC-Seg achieved the highest validation IoU (peak 78.8\%, final 78.7\%). CMC-Seg also had the lowest loss among the benchmark baselines (final 0.231, minimum 0.229 at epoch 81) and reached 90\% of its maximum IoU by epoch 21, faster than the Transformer baselines (nnFormer: epoch 24; UNETR: epoch 34) and comparable to the CNN baselines (3D U-Net: epoch 20; SegResNet: epoch 29).

\begin{table*}[!t]
	\centering
	\caption{Dice score ($\pm$ standard deviation) of tumor classes and subregions on the test data.}
	\label{tab:dice_scores}
	\newcommand{\bv}[2]{\underline{\textbf{#1}}\newline\underline{\textbf{$\pm$\,#2}}}
	\newcommand{\sv}[2]{\textbf{#1}\newline\textbf{$\pm$\,#2}}
	\newcommand{\nv}[2]{#1\newline$\pm$\,#2}
	\footnotesize
	\setlength{\tabcolsep}{4pt}
	\renewcommand{\arraystretch}{1.35}
	\begin{tabularx}{\textwidth}{@{}>{\RaggedRight\arraybackslash}p{2.7cm}
			*{7}{>{\centering\arraybackslash}X}@{}}
		\toprule[1.2pt]
		\rowcolor{gray!25}
		\textbf{Model} & \textbf{WT} & \textbf{TC} & \textbf{ET} &
		\textbf{NETC} & \textbf{SNFH} & \textbf{RC} & \textbf{Mean} \\
		\midrule[0.8pt]
		3D U-Net\newline(\cite{cciccek20163d}) &
		\sv{91.44}{9.17}   & \nv{81.58}{26.08} & \nv{79.93}{26.40} &
		\sv{76.49}{32.06}  & \sv{87.82}{11.57} & \bv{79.83}{28.90} &
		\sv{81.10}{14.62} \\
		\addlinespace[3pt]
		\rowcolor{gray!8}
		SegResNet\newline(\cite{myronenko20183d}) &
		\bv{91.48}{9.14}   & \nv{82.16}{24.91} & \sv{81.27}{24.63} &
		\nv{75.28}{33.95}  & \nv{87.78}{11.43} & \nv{78.30}{29.37} &
		\nv{80.89}{14.23} \\
		\addlinespace[3pt]
		Swin-UNETR\newline(\cite{hatamizadeh2021swin}) &
		\nv{90.98}{9.33}   & \nv{81.26}{25.85} & \nv{80.56}{25.41} &
		\nv{73.93}{33.88}  & \nv{87.23}{11.75} & \nv{76.45}{30.56} &
		\nv{79.98}{15.50} \\
		\addlinespace[3pt]
		\rowcolor{gray!8}
		UNETR\newline(\cite{hatamizadeh2022unetr}) &
		\nv{90.04}{9.98}   & \nv{79.14}{27.39} & \nv{77.78}{27.78} &
		\nv{71.93}{34.87}  & \nv{85.93}{12.59} & \nv{70.60}{34.76} &
		\nv{77.37}{16.05} \\
		\addlinespace[3pt]
		nnFormer\newline(\cite{zhou2023nnformer}) &
		\nv{89.44}{9.80}   & \nv{79.27}{27.21} & \nv{77.40}{26.48} &
		\nv{73.85}{32.59}  & \nv{85.15}{12.20} & \nv{73.84}{31.30} &
		\nv{78.07}{14.64} \\
		\addlinespace[3pt]
		\rowcolor{gray!8}
		SegMamba\newline(\cite{xing2024segmambalongrangesequentialmodeling}) &
		\sv{91.44}{9.47}   & \bv{82.64}{24.55} & \bv{81.34}{24.75} &
		\nv{74.94}{33.27}  & \bv{87.99}{11.98} & \nv{77.57}{30.16} &
		\nv{80.82}{15.02} \\
		\addlinespace[3pt]
		CMC-Seg &
		\nv{91.31}{9.27}   & \sv{82.39}{24.49} & \nv{80.73}{25.28} &
		\bv{77.09}{31.79}  & \nv{87.53}{12.04} & \sv{78.63}{29.91} &
		\bv{81.20}{14.37} \\
		\bottomrule[1.2pt]
	\end{tabularx}
	
	\vspace{3pt}
	\parbox{\textwidth}{\footnotesize
		Higher Dice score is better ($\uparrow$). The best-performing values are \underline{\textbf{bold and underlined}}; the second-best values are \textbf{bold} (tied second-best values are all marked).
		WT, whole tumor; TC, tumor core; NETC, non-enhancing tumor core; SNFH, surrounding non-enhancing FLAIR hyperintensity; ET, enhancing tumor; RC, resection cavity.}
\end{table*}

Table~3 provides the entire test-dataset Dice score results. In both mean Dice score and NETC Dice score, CMC-Seg achieves the highest reported overall mean Dice and NETC Dice scores, although its performance advantage varies across tumor regions and treatment subgroups. Specifically, CMC-Seg achieved a mean Dice score of 81.20\% $\pm$ 14.37\% and an NETC Dice score of 77.09\% $\pm$ 31.79\%, with the improvement over Transformer-based baselines (SwinUNETR, UNETR, nnFormer) reaching statistical significance (FDR-adjusted $P < 0.05$). CMC-Seg also achieved RC and TC Dice scores of 78.63\% $\pm$ 29.91\% and 82.39\% $\pm$ 24.49\%, comparable to 3D U-Net, SegResNet, and SegMamba, and statistically significant over Transformer-based baselines (FDR-adjusted $P < 0.05$). 

\begin{table*}[!t]
	\centering
	\caption{Dice score ($\pm$ standard deviation) of tumor classes and subregions based on treatment status on the test data.}
	\label{tab:dice_treatment}
	\footnotesize
	\setlength{\tabcolsep}{4pt}
	\renewcommand{\arraystretch}{1.3}
	\begin{tabularx}{\textwidth}{@{}>{\RaggedRight\arraybackslash}p{2.6cm}
			>{\RaggedRight\arraybackslash}p{1.9cm}
			*{5}{>{\centering\arraybackslash}X}@{}}
		\toprule[1.2pt]
		\rowcolor{gray!25}
		\textbf{Model} & \textbf{Subgroup} & \textbf{WT} & \textbf{TC} & \textbf{ET} &
		\textbf{RC} & \textbf{Mean} \\
		\midrule[0.8pt]
		3D U-Net\newline(\cite{cciccek20163d}) & Pre-treatment &
		92.0\newline$\pm$\,8.5  & 88.1\newline$\pm$\,17.3 & 84.0\newline$\pm$\,19.6 &
		---                      & 80.5\newline$\pm$\,15.2 \\
		\rowcolor{gray!8}
		& Post-treatment &
		91.0\newline$\pm$\,9.6  & 76.7\newline$\pm$\,30.2 & 76.9\newline$\pm$\,30.1 &
		79.8\newline$\pm$\,28.9 & 81.5\newline$\pm$\,14.2 \\
		\addlinespace[3pt]
		SegResNet\newline(\cite{myronenko20183d}) & Pre-treatment &
		92.2\newline$\pm$\,8.1  & 87.2\newline$\pm$\,18.3 & 85.0\newline$\pm$\,17.8 &
		---                      & 80.5\newline$\pm$\,15.4 \\
		\rowcolor{gray!8}
		& Post-treatment &
		91.0\newline$\pm$\,9.8  & 78.4\newline$\pm$\,28.3 & 78.4\newline$\pm$\,28.4 &
		78.3\newline$\pm$\,29.4 & 81.2\newline$\pm$\,13.3 \\
		\addlinespace[3pt]
		Swin-UNETR\newline(\cite{hatamizadeh2021swin}) & Pre-treatment &
		91.8\newline$\pm$\,8.5  & 87.8\newline$\pm$\,15.7 & 84.4\newline$\pm$\,17.5 &
		---                      & 80.5\newline$\pm$\,14.8 \\
		\rowcolor{gray!8}
		& Post-treatment &
		90.4\newline$\pm$\,9.9  & 76.4\newline$\pm$\,30.5 & 77.7\newline$\pm$\,29.7 &
		76.5\newline$\pm$\,30.6 & 79.6\newline$\pm$\,16.0 \\
		\addlinespace[3pt]
		UNETR\newline(\cite{hatamizadeh2022unetr}) & Pre-treatment &
		91.3\newline$\pm$\,9.1  & 86.3\newline$\pm$\,17.7 & 82.9\newline$\pm$\,20.0 &
		---                      & 78.2\newline$\pm$\,16.2 \\
		\rowcolor{gray!8}
		& Post-treatment &
		89.1\newline$\pm$\,10.5 & 73.7\newline$\pm$\,31.8 & 74.0\newline$\pm$\,31.9 &
		70.6\newline$\pm$\,34.8 & 76.8\newline$\pm$\,15.9 \\
		\addlinespace[3pt]
		nnFormer\newline(\cite{zhou2023nnformer}) & Pre-treatment &
		90.8\newline$\pm$\,8.7  & 86.7\newline$\pm$\,17.6 & 82.8\newline$\pm$\,16.8 &
		---                      & 78.6\newline$\pm$\,14.7 \\
		\rowcolor{gray!8}
		& Post-treatment &
		88.4\newline$\pm$\,10.4 & 73.7\newline$\pm$\,31.5 & 73.4\newline$\pm$\,31.2 &
		73.8\newline$\pm$\,31.3 & 77.7\newline$\pm$\,14.6 \\
		\addlinespace[3pt]
		SegMamba\newline(\cite{xing2024segmambalongrangesequentialmodeling}) & Pre-treatment &
		92.0\newline$\pm$\,8.9  & 88.5\newline$\pm$\,15.0 & 85.0\newline$\pm$\,17.2 &
		---                      & 80.9\newline$\pm$\,15.0 \\
		\rowcolor{gray!8}
		& Post-treatment &
		91.0\newline$\pm$\,9.8  & 78.2\newline$\pm$\,29.0 & 78.6\newline$\pm$\,28.9 &
		77.6\newline$\pm$\,30.2 & 80.8\newline$\pm$\,15.0 \\
		\addlinespace[3pt]
		CMC-Seg & Pre-treatment &
		91.9\newline$\pm$\,8.5  & 87.3\newline$\pm$\,16.9 & 83.3\newline$\pm$\,20.1 &
		---                      & 80.5\newline$\pm$\,15.5 \\
		\rowcolor{gray!8}
		& Post-treatment &
		90.9\newline$\pm$\,9.8  & 78.7\newline$\pm$\,28.4 & 78.8\newline$\pm$\,28.4 &
		78.6\newline$\pm$\,29.9 & 81.7\newline$\pm$\,13.4 \\
		\bottomrule[1.2pt]
	\end{tabularx}
	
	\vspace{3pt}
	\parbox{\textwidth}{\footnotesize
		Higher Dice score is better ($\uparrow$). ``---'' indicates that the subregion is not defined for the corresponding treatment status (no resection cavity in pre-treatment cases). WT, whole tumor; TC, tumor core; ET, enhancing tumor; RC, resection cavity.}
\end{table*}

Table~4 stratifies Dice score by treatment status. In the pre-treatment subgroup, CMC-Seg showed a mean Dice score of 80.5\% $\pm$ 15.5\%, not significantly different from 3D U-Net, SegResNet, and SwinUNETR, and marginally below SegMamba (80.9\% $\pm$ 15.0\%), while remaining significantly ahead of nnFormer and UNETR (FDR-adjusted $P < 0.05$). In the post-treatment subgroup, CMC-Seg demonstrated the highest mean Dice score (81.7\% $\pm$ 13.4\%) among benchmark models, with the margin over Transformer-based models reaching significance (FDR-adjusted $P < 0.05$). This indicates that CMC-Seg's segmentation advantage is most pronounced in post-treatment cases, while it remained competitive with the strongest CNN baselines on pre-treatment scans. 

At the sub-region level, in the pre-treatment subgroup, CMC-Seg's WT (91.9\% $\pm$ 8.5\%), TC (87.3\% $\pm$ 16.9\%), and ET (83.3\% $\pm$ 20.1\%) Dice scores were competitive with the leading CNN- and Mamba-based baselines. SegResNet attained the highest WT score (92.2\% $\pm$ 8.1\%), SegMamba led on TC (88.5\% $\pm$ 15.0\%), and SegResNet and SegMamba jointly led on ET (85.0\% $\pm$ 17.8\% and 85.0\% $\pm$ 17.2\%, respectively). CMC-Seg nonetheless remains clearly superior to nnFormer and UNETR across all three regions (FDR-adjusted $P < 0.05$). 

\begin{figure*}[!t]
	\centering
	\includegraphics[width=\textwidth,height=0.88\textheight,keepaspectratio]{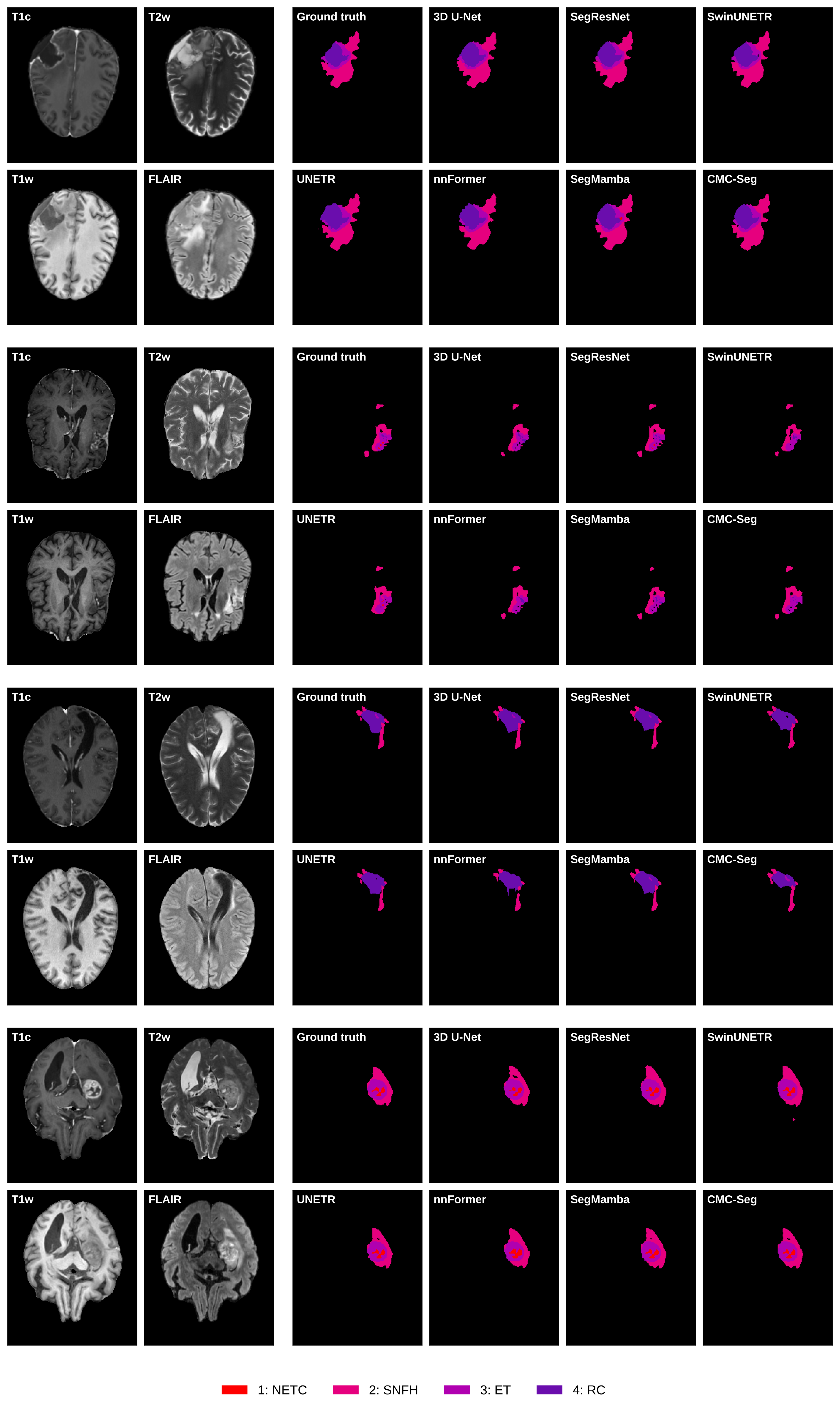}
	\caption{Visualization of the segmentation results of CMC-Seg and benchmark models on the BraTS 2025 dataset.}
	\label{FIG:4}
\end{figure*}

In the post-treatment subgroup, CMC-Seg achieved the highest TC (78.7\% $\pm$ 28.4\%) and ET (78.8\% $\pm$ 28.4\%) Dice scores, slightly higher than SegResNet (78.4\% $\pm$ 28.3\% / 78.4\% $\pm$ 28.4\%) and SegMamba (78.2\% $\pm$ 29.0\% / 78.6\% $\pm$ 28.9\%), while its WT score (90.9\% $\pm$ 9.8\%) was slightly lower than the three strongest baselines (3D U-Net, 91.0\% $\pm$ 9.6\%; SegMamba, 91.0\% $\pm$ 9.8\%; SegResNet, 91.0\% $\pm$ 9.8\%). These findings indicate that CMC-Seg maintains competitive performance in pre-treatment cases while demonstrating more favorable relative performance in the post-treatment subgroup, particularly for TC, ET, and RC segmentation. Thus, CMC-Seg's segmentation accuracy is mainly focused on the post-treatment subgroup. Fig.~4 shows segmentation results of CMC-Seg and benchmark models on the test dataset.

\begin{table*}[!t]
	\centering
	\caption{HD95 score (mm; lower is better) $\pm$ standard deviation of tumor classes and subregions on the test data.}
	\label{tab:hd95_scores}
	\footnotesize
	\setlength{\tabcolsep}{3pt}
	\renewcommand{\arraystretch}{1.35}
	\begin{tabularx}{\textwidth}{@{}>{\RaggedRight\arraybackslash}p{2.6cm}
			*{7}{>{\centering\arraybackslash}X}@{}}
		\toprule[1.2pt]
		\rowcolor{gray!25}
		\textbf{Model} & \textbf{WT} & \textbf{TC} & \textbf{ET} &
		\textbf{NETC} & \textbf{SNFH} & \textbf{RC} & \textbf{Mean} \\
		\midrule[0.8pt]
		3D U-Net\newline(\cite{cciccek20163d}) &
		\underline{\textbf{5.00}}\newline\underline{\textbf{$\pm$\,8.90}} &
		\underline{\textbf{6.02}}\newline\underline{\textbf{$\pm$\,12.51}} &
		\textbf{5.44}\newline\textbf{$\pm$\,12.28} &
		5.11\newline$\pm$\,9.62 &
		\underline{\textbf{4.82}}\newline\underline{\textbf{$\pm$\,8.06}} &
		\textbf{4.51}\newline\textbf{$\pm$\,7.91} &
		\textbf{5.15}\newline\textbf{$\pm$\,5.87} \\
		\addlinespace[3pt]
		\rowcolor{gray!8}
		SegResNet\newline(\cite{myronenko20183d}) &
		\textbf{5.08}\newline\textbf{$\pm$\,9.48} &
		\textbf{6.19}\newline\textbf{$\pm$\,11.83} &
		\underline{\textbf{5.34}}\newline\underline{\textbf{$\pm$\,11.56}} &
		\underline{\textbf{4.75}}\newline\underline{\textbf{$\pm$\,9.12}} &
		\textbf{4.89}\newline\textbf{$\pm$\,8.71} &
		4.69\newline$\pm$\,7.40 &
		\underline{\textbf{5.03}}\newline\underline{\textbf{$\pm$\,5.75}} \\
		\addlinespace[3pt]
		Swin-UNETR\newline(\cite{hatamizadeh2021swin}) &
		6.10\newline$\pm$\,10.10 &
		7.09\newline$\pm$\,13.98 &
		5.84\newline$\pm$\,12.60 &
		5.75\newline$\pm$\,11.45 &
		5.73\newline$\pm$\,9.16 &
		5.30\newline$\pm$\,8.11 &
		5.81\newline$\pm$\,6.52 \\
		\addlinespace[3pt]
		\rowcolor{gray!8}
		UNETR\newline(\cite{hatamizadeh2022unetr}) &
		6.34\newline$\pm$\,10.01 &
		7.36\newline$\pm$\,12.83 &
		6.25\newline$\pm$\,12.52 &
		5.95\newline$\pm$\,10.51 &
		6.20\newline$\pm$\,9.46 &
		6.06\newline$\pm$\,8.90 &
		6.21\newline$\pm$\,6.21 \\
		\addlinespace[3pt]
		nnFormer\newline(\cite{zhou2023nnformer}) &
		5.61\newline$\pm$\,10.07 &
		6.97\newline$\pm$\,14.12 &
		6.11\newline$\pm$\,13.84 &
		4.99\newline$\pm$\,9.64 &
		5.46\newline$\pm$\,9.39 &
		5.25\newline$\pm$\,7.86 &
		5.55\newline$\pm$\,6.76 \\
		\addlinespace[3pt]
		\rowcolor{gray!8}
		SegMamba\newline(\cite{xing2024segmambalongrangesequentialmodeling}) &
		5.60\newline$\pm$\,10.17 &
		7.06\newline$\pm$\,14.06 &
		6.16\newline$\pm$\,13.51 &
		5.60\newline$\pm$\,11.67 &
		5.32\newline$\pm$\,9.42 &
		4.94\newline$\pm$\,8.72 &
		5.68\newline$\pm$\,7.30 \\
		\addlinespace[3pt]
		CMC-Seg &
		5.19\newline$\pm$\,9.72 &
		6.31\newline$\pm$\,12.09 &
		5.65\newline$\pm$\,11.91 &
		\textbf{4.90}\newline\textbf{$\pm$\,9.30} &
		5.15\newline$\pm$\,9.06 &
		\underline{\textbf{4.39}}\newline\underline{\textbf{$\pm$\,7.24}} &
		5.20\newline$\pm$\,5.83 \\
		\bottomrule[1.2pt]
	\end{tabularx}
	
	\vspace{3pt}
	\parbox{\textwidth}{\footnotesize
		Lower HD95 is better ($\downarrow$). The best-performing values are \underline{\textbf{bold and underlined}}; the second-best values are \textbf{bold}.\newline
		WT, whole tumor; TC, tumor core; NETC, non-enhancing tumor core; SNFH, surrounding non-enhancing FLAIR hyperintensity; ET, enhancing tumor; RC, resection cavity.}
\end{table*}

Table~5 reports the HD95 score for all tumor classes and subregions. CMC-Seg achieved an HD95 score of 4.39 $\pm$ 7.24 mm for the RC class, the lowest RC boundary error among all seven models evaluated. CMC-Seg's HD95 score did not differ significantly from the two strongest CNN-based baselines, 3D U-Net (mean HD95 score 5.15 mm) and SegResNet (5.03 mm), for the mean and WT HD95 scores. Compared with 3D U-Net, two subregion comparisons were significant in 3D U-Net's favor, including TC and ET (FDR-adjusted $P < 0.05$), indicating a modest boundary-precision advantage. Additionally, CMC-Seg showed lower HD95 scores than SegMamba across all subregions and classes; we found no significant differences on any HD95 metric. 

\begin{table*}[!t]
	\centering
	\caption{HD95 score (lower is better) $\pm$ std of tumor classes and subregions based on treatment status on the test data.}
	\label{tab:hd95_treatment}
	\footnotesize
	\setlength{\tabcolsep}{4pt}
	\renewcommand{\arraystretch}{1.35}
	\begin{tabularx}{\textwidth}{@{}>{\RaggedRight\arraybackslash}p{2.6cm}
			>{\RaggedRight\arraybackslash}p{1.9cm}
			*{5}{>{\centering\arraybackslash}X}@{}}
		\toprule[1.2pt]
		\rowcolor{gray!25}
		\textbf{Model} & \textbf{Subgroup} & \textbf{WT} & \textbf{TC} & \textbf{ET} &
		\textbf{RC} & \textbf{Mean} \\
		\midrule[0.8pt]
		3D U-Net\newline(\cite{cciccek20163d}) & Pre-treatment &
		5.90\newline$\pm$\,9.02  & 5.71\newline$\pm$\,12.07 & 4.98\newline$\pm$\,11.34 &
		---                       & 5.95\newline$\pm$\,6.23 \\
		\rowcolor{gray!8}
		& Post-treatment &
		4.32\newline$\pm$\,8.75  & 6.28\newline$\pm$\,12.86 & 5.81\newline$\pm$\,12.96 &
		4.51\newline$\pm$\,7.91  & 4.55\newline$\pm$\,5.52 \\
		\addlinespace[3pt]
		SegResNet\newline(\cite{myronenko20183d}) & Pre-treatment &
		6.26\newline$\pm$\,9.85  & 6.11\newline$\pm$\,11.11 & 4.85\newline$\pm$\,10.46 &
		---                       & 5.93\newline$\pm$\,6.14 \\
		\rowcolor{gray!8}
		& Post-treatment &
		4.18\newline$\pm$\,9.09  & 6.26\newline$\pm$\,12.37 & 5.72\newline$\pm$\,12.36 &
		4.69\newline$\pm$\,7.40  & 4.35\newline$\pm$\,5.34 \\
		\addlinespace[3pt]
		Swin-UNETR\newline(\cite{hatamizadeh2021swin}) & Pre-treatment &
		7.05\newline$\pm$\,10.19 & 5.68\newline$\pm$\,10.02 & 4.48\newline$\pm$\,9.42 &
		---                       & 6.47\newline$\pm$\,7.14 \\
		\rowcolor{gray!8}
		& Post-treatment &
		5.38\newline$\pm$\,9.96  & 8.23\newline$\pm$\,16.42 & 6.91\newline$\pm$\,14.54 &
		5.30\newline$\pm$\,8.11  & 5.32\newline$\pm$\,5.96 \\
		\addlinespace[3pt]
		UNETR\newline(\cite{hatamizadeh2022unetr}) & Pre-treatment &
		6.67\newline$\pm$\,9.87  & 6.75\newline$\pm$\,10.66 & 4.92\newline$\pm$\,9.35 &
		---                       & 6.55\newline$\pm$\,5.90 \\
		\rowcolor{gray!8}
		& Post-treatment &
		6.09\newline$\pm$\,10.11 & 7.87\newline$\pm$\,14.38 & 7.34\newline$\pm$\,14.53 &
		6.06\newline$\pm$\,8.90  & 5.96\newline$\pm$\,6.42 \\
		\addlinespace[3pt]
		nnFormer\newline(\cite{zhou2023nnformer}) & Pre-treatment &
		6.50\newline$\pm$\,10.51 & 6.30\newline$\pm$\,11.84 & 4.86\newline$\pm$\,10.79 &
		---                       & 6.03\newline$\pm$\,6.85 \\
		\rowcolor{gray!8}
		& Post-treatment &
		4.95\newline$\pm$\,9.67  & 7.52\newline$\pm$\,15.75 & 7.13\newline$\pm$\,15.85 &
		5.25\newline$\pm$\,7.86  & 5.19\newline$\pm$\,6.67 \\
		\addlinespace[3pt]
		SegMamba\newline(\cite{xing2024segmambalongrangesequentialmodeling}) & Pre-treatment &
		6.68\newline$\pm$\,10.96 & 6.28\newline$\pm$\,12.65 & 4.91\newline$\pm$\,11.13 &
		---                       & 6.36\newline$\pm$\,8.53 \\
		\rowcolor{gray!8}
		& Post-treatment &
		4.80\newline$\pm$\,9.45  & 7.69\newline$\pm$\,15.08 & 7.16\newline$\pm$\,15.06 &
		4.94\newline$\pm$\,8.72  & 5.18\newline$\pm$\,6.18 \\
		\addlinespace[3pt]
		CMC-Seg & Pre-treatment &
		6.17\newline$\pm$\,10.88 & 6.09\newline$\pm$\,11.20 & 5.18\newline$\pm$\,10.77 &
		---                       & 6.13\newline$\pm$\,6.23 \\
		\rowcolor{gray!8}
		& Post-treatment &
		4.45\newline$\pm$\,8.68  & 6.48\newline$\pm$\,12.75 & 6.02\newline$\pm$\,12.73 &
		4.39\newline$\pm$\,7.24  & 4.50\newline$\pm$\,5.40 \\
		\bottomrule[1.2pt]
	\end{tabularx}
	
	\vspace{3pt}
	\parbox{\textwidth}{\footnotesize
		Lower HD95 is better ($\downarrow$). ``---'' indicates that the subregion is not defined for the corresponding treatment status (no resection cavity in pre-treatment cases).
		WT, whole tumor; TC, tumor core; ET, enhancing tumor; RC, resection cavity.}
\end{table*}

In contrast, CMC-Seg showed a clear and consistent boundary-accuracy advantage over the Transformer-based baselines. The mean HD95 score was significantly lower than nnFormer, SwinUNETR, and UNETR (FDR-adjusted $P < 0.05$). This pattern held across all classes and subregions individually, with UNETR demonstrating the most pronounced deficit (RC: mean difference $-1.67$ mm, $P = 2.6 \times 10^{-8}$). In the pre-treatment subgroup (Table~6), CMC-Seg's mean boundary error (6.13 $\pm$ 6.23 mm) is comparable to 3D U-Net (5.95 $\pm$ 6.23 mm), SegResNet (5.93 $\pm$ 6.14 mm), and nnFormer (6.03 $\pm$ 6.85 mm), and lower than SegMamba (6.36 $\pm$ 8.53 mm), SwinUNETR (6.47 $\pm$ 7.14 mm), and UNETR (6.55 $\pm$ 5.90 mm). 

In the post-treatment subgroup, CMC-Seg's mean HD95 score dropped to 4.50 $\pm$ 5.40 mm, the second-lowest of all models, trailing only SegResNet (4.35 $\pm$ 5.34 mm). The regional analysis revealed a more heterogeneous picture compared to the mean HD95 values. In the pre-treatment subgroup, CMC-Seg's WT HD95 score (6.17 $\pm$ 10.88 mm) was the second-lowest of all seven models, behind only 3D U-Net (5.90 mm). The ET was the one region where CMC-Seg underperformed in the pre-treatment subgroup, posting the highest boundary error of any model at 5.18 $\pm$ 10.77 mm. In the post-treatment subgroup, the results shifted notably: CMC-Seg's ET HD95 score improved to 6.02 $\pm$ 12.73 mm, which is significantly lower than nnFormer and UNETR, as well as its TC HD95 score (6.48 $\pm$ 12.75 mm) was significantly lower than Transformer baselines. Taken together, CMC-Seg displays competitive boundary precision results in the pre-treatment subgroup, while these results improve in the post-treatment subgroup, particularly compared to Transformer baselines.

\begin{table*}[!t]
	\centering
	\caption{Precision and recall ($\pm$ standard deviation) of tumor classes and subregions on the test data.}
	\label{tab:precision_recall}
	\footnotesize
	\setlength{\tabcolsep}{3pt}
	\renewcommand{\arraystretch}{1.35}
	\begin{tabularx}{\textwidth}{@{}>{\RaggedRight\arraybackslash}p{2.3cm}
			>{\RaggedRight\arraybackslash}p{1.9cm}
			*{7}{>{\centering\arraybackslash}X}@{}}
		\toprule[1.2pt]
		\rowcolor{gray!25}
		\textbf{Metric} & \textbf{Subgroup} & \textbf{NETC} & \textbf{SNFH} & \textbf{ET} &
		\textbf{WT} & \textbf{RC} & \textbf{TC} & \textbf{Mean} \\
		\midrule[0.8pt]
		Precision (\%)\newline(Pre.) & Pre-treatment &
		79.6\newline$\pm$\,25.5  & 83.0\newline$\pm$\,14.3 & 84.2\newline$\pm$\,20.7 &
		90.5\newline$\pm$\,9.7   & ---                      & 88.6\newline$\pm$\,17.7 &
		82.3\newline$\pm$\,13.8 \\
		\rowcolor{gray!8}
		& Post-treatment &
		90.0\newline$\pm$\,21.1  & 88.4\newline$\pm$\,12.8 & 83.3\newline$\pm$\,24.4 &
		89.4\newline$\pm$\,12.0  & 84.1\newline$\pm$\,24.0  & 83.3\newline$\pm$\,24.7 &
		86.5\newline$\pm$\,10.6 \\
		\addlinespace[3pt]
		Recall (\%)\newline(Sensitivity) & Pre-treatment &
		77.4\newline$\pm$\,29.1  & 88.0\newline$\pm$\,13.5 & 87.6\newline$\pm$\,15.6 &
		94.1\newline$\pm$\,9.2   & ---                      & 89.8\newline$\pm$\,15.3 &
		84.3\newline$\pm$\,13.3 \\
		\rowcolor{gray!8}
		& Post-treatment &
		81.1\newline$\pm$\,32.4  & 92.1\newline$\pm$\,9.7  & 82.7\newline$\pm$\,24.9 &
		93.2\newline$\pm$\,9.1   & 83.2\newline$\pm$\,26.8  & 82.7\newline$\pm$\,25.1 &
		84.9\newline$\pm$\,12.8 \\
		\bottomrule[1.2pt]
	\end{tabularx}
	
	\vspace{3pt}
	\parbox{\textwidth}{\footnotesize
		Recall is also reported as sensitivity. ``---'' indicates that the subregion is not defined for the corresponding treatment status (no resection cavity in pre-treatment cases).
		WT, whole tumor; TC, tumor core; NETC, non-enhancing tumor core; SNFH, surrounding non-enhancing FLAIR hyperintensity; ET, enhancing tumor; RC, resection cavity.}
\end{table*}

To characterize the trade-off between over- and under-segmentation, we additionally report precision and recall (sensitivity), stratified by pre- and post-treatment subgroup (Table~7), as region composition and boundary complexity differ notably between these two clinical subgroups. In the pre-treatment subgroup, CMC-Seg achieves a mean precision of 82.3\% $\pm$ 13.8\% and mean recall of 84.3\% $\pm$ 13.3\%. Recall exceeds precision across all pre-treatment regions except NETC, a reasonable bias for a segmentation task where under-detection of tumor tissue carries greater clinical risk than modest over-segmentation. In the post-treatment subgroup, which includes the additional RC region and more heterogeneous tissue due to treatment, mean precision rose to 86.5\% $\pm$ 10.6\%, while mean recall was 84.9\% $\pm$ 12.8\%. RC also achieves precision of 84.1\% $\pm$ 24\% and recall of 83.2\% $\pm$ 26.8\%, reflecting reasonably symmetric performance in this treatment-cavity region despite its irregular, heterogeneous morphology. Across both subgroups, SNFH shows the strongest recall (88\% -- 92.1\%), consistent with its Dice score being the most accurate tumor class, while NETC shows the largest precision--recall gap in the post-treatment subgroup (90\% precision vs. 81.1\% recall), suggesting the CMC-Seg is more conservative (higher precision, lower recall) when delineating NETC after treatment-induced tissue changes.

\subsection{Computational complexity analysis}
We examined the computational cost of achieving this accuracy (Table~8). The model's parameters, GFLOPs, and average inference time are computed at inference with float 32 and batch size 1. CMC-Seg has 46.76M parameters, positioning it between the CNN baselines (3D U-Net, 16.32M; SegResNet, 25.21M) and the Transformer models (UNETR, 143.16M; nnFormer, 150.06M).

CMC-Seg required the fewest inference GFLOPs of any model evaluated (381.49 GFLOPs), lower than lightweight models like nnFormer (494.86 GFLOPs), and showed a large gap compared with UNETR (4123.45 GFLOPs). CMC-Seg peak training VRAM usage was moderate at 7.48 GB, comparable to Swin-UNETR (7.57 GB) and below SegResNet, UNETR, and SegMamba. This favorable accuracy-per-FLOP profile did not translate into faster wall-clock inference: CMC-Seg's inference time of 1883.27 $\pm$ 19.68 ms was the slowest among all benchmark baselines. This reflects a known characteristic of Mamba-based SSMs, whose sequential scan operations are memory-bound rather than compute-bound, so low FLOPs do not translate proportionally into low latency on current hardware. 

\begin{table*}[!t]
	\centering
	\caption{Computational efficiency and complexity metrics of the models.}
	\label{tab:complexity}
	\footnotesize
	\setlength{\tabcolsep}{4pt}
	\renewcommand{\arraystretch}{1.3}
	\begin{tabularx}{\textwidth}{@{}>{\RaggedRight\arraybackslash}p{3.0cm}
			*{4}{>{\centering\arraybackslash}X}@{}}
		\toprule[1.2pt]
		\rowcolor{gray!25}
		\textbf{Model} & \textbf{Parameters} & \textbf{GFLOPs} & \textbf{VRAM (GB)} &
		\textbf{Inference Time (ms)} \\
		\midrule[0.8pt]
		3D U-Net\newline(\cite{cciccek20163d}) &
		\underline{\textbf{16.32M}} & 3807.41 & 7.81 & 1089.00\,$\pm$\,10.92 \\
		\addlinespace[2pt]
		\rowcolor{gray!8}
		SegResNet\newline(\cite{myronenko20183d}) &
		\textbf{25.21M} & 1174.14 & 8.50 &
		\underline{\textbf{411.41\,$\pm$\,2.41}} \\
		\addlinespace[2pt]
		Swin-UNETR\newline(\cite{hatamizadeh2021swin}) &
		62.19M & 1550.16 & 7.57 & \textbf{929.35\,$\pm$\,6.50} \\
		\addlinespace[2pt]
		\rowcolor{gray!8}
		UNETR\newline(\cite{hatamizadeh2022unetr}) &
		143.16M & 4123.45 & 8.43 & 1154.32\,$\pm$\,9.14 \\
		\addlinespace[2pt]
		nnFormer\newline(\cite{zhou2023nnformer}) &
		150.06M & \textbf{494.86} & \underline{\textbf{6.29}} & 1091.87\,$\pm$\,6.59 \\
		\addlinespace[2pt]
		\rowcolor{gray!8}
		SegMamba\newline(\cite{xing2024segmambalongrangesequentialmodeling}) &
		67.42M & 2928.91 & 8.35 & 1392.28\,$\pm$\,12.21 \\
		\addlinespace[2pt]
		CMC-Seg &
		46.76M & \underline{\textbf{381.49}} & \textbf{7.48} & 1883.27\,$\pm$\,19.68 \\
		\bottomrule[1.2pt]
	\end{tabularx}
	
	\vspace{3pt}
	\parbox{\textwidth}{\footnotesize
		Lower values are better for all reported metrics. The best-performing values are \underline{\textbf{bold and underlined}}; the second-best values are \textbf{bold}.
		GFLOPs, giga floating-point operations; VRAM, video random-access memory.}
\end{table*}

\begin{figure*}
	\centering
	\includegraphics[width=\textwidth]{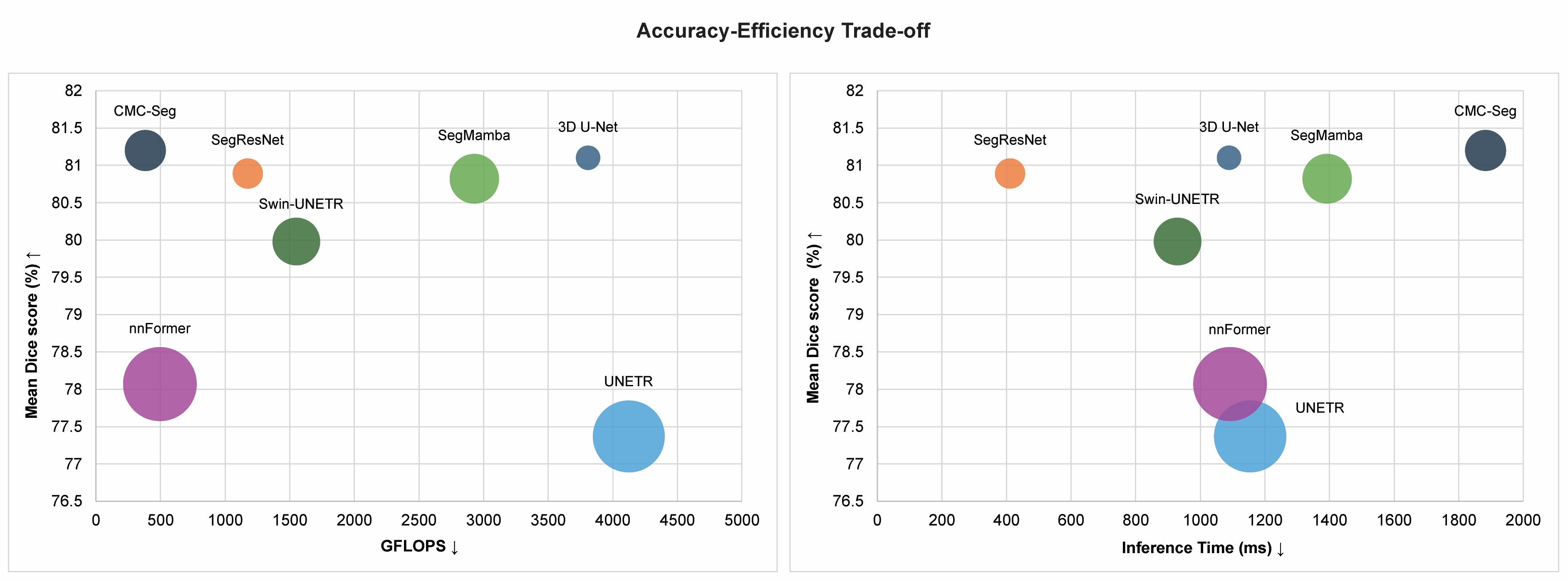}
	\caption{The computational cost versus accuracy trade-off. Model parameters are shown by their circle size. CMC-Seg achieves the best mean Dice score-to-GFLOPs ratio, although it demonstrates the slowest inference time.}
	\label{FIG:5}
\end{figure*}

Taken together, these results position CMC-Seg as achieving the best aggregate segmentation performance at the lowest computational cost in FLOPs, by sacrificing inference time relative to CNN- and Transformer-based alternatives (Fig.~5).

\subsection{Ablation studies}
To validate our architectural design and analyze each key component, this section presents a series of ablation studies (Table~9). Starting with a basic Mamba-CNN baseline, we gradually added our proposed modules and measured the performance enhancement at each stage. In the first analysis, we investigated the specific design of our MDM block. Model (B) replaces our proposed direction-aware scanning mechanism with a simple Mamba block that flattens the 3D feature map and processes it as a single 1D sequence. 

\begin{table*}[!t]
	\centering
	\caption{Dice and HD95 score ($\pm$ standard deviation) of the ablation models.}
	\label{tab:ablation}
	\footnotesize
	\setlength{\tabcolsep}{2pt}
	\renewcommand{\arraystretch}{1.3}
	\begin{tabularx}{\textwidth}{@{}>{\RaggedRight\arraybackslash}p{2.3cm}
			*{10}{>{\centering\arraybackslash}X}@{}}
		\toprule[1.2pt]
		\rowcolor{gray!25}
		& \multicolumn{5}{c}{\textbf{Dice score (\%)} $\uparrow$}
		& \multicolumn{5}{c}{\textbf{HD95 score (mm)} $\downarrow$} \\
		\cmidrule(lr){2-6}\cmidrule(l){7-11}
		\rowcolor{gray!25}
		\textbf{Model} & \textbf{NETC} & \textbf{SNFH} & \textbf{ET} & \textbf{RC} & \textbf{Mean}
		& \textbf{NETC} & \textbf{SNFH} & \textbf{ET} & \textbf{RC} & \textbf{Mean} \\
		\midrule[0.8pt]
		Model (B) &
		74.50\newline$\pm$\,33.99 & 87.58\newline$\pm$\,11.78 & 80.33\newline$\pm$\,25.87 &
		75.17\newline$\pm$\,32.00 & 79.93\newline$\pm$\,15.04 &
		5.41\newline$\pm$\,10.30  & 4.92\newline$\pm$\,8.80  & 5.65\newline$\pm$\,12.44 &
		4.78\newline$\pm$\,7.19   & 5.28\newline$\pm$\,6.08 \\
		\addlinespace[2pt]
		\rowcolor{gray!8}
		Model (C) &
		75.73\newline$\pm$\,33.00 & 87.75\newline$\pm$\,11.54 & 80.96\newline$\pm$\,25.59 &
		78.48\newline$\pm$\,29.52 & 81.00\newline$\pm$\,14.45 &
		4.79\newline$\pm$\,8.95   & 5.15\newline$\pm$\,9.17  & 5.77\newline$\pm$\,12.72 &
		4.90\newline$\pm$\,7.53   & 5.27\newline$\pm$\,6.03 \\
		\addlinespace[2pt]
		Model (D) &
		75.24\newline$\pm$\,33.00 & 87.72\newline$\pm$\,11.88 & 80.38\newline$\pm$\,25.51 &
		76.44\newline$\pm$\,31.15 & 80.39\newline$\pm$\,14.52 &
		5.38\newline$\pm$\,10.32  & 5.18\newline$\pm$\,9.16  & 5.88\newline$\pm$\,13.43 &
		4.72\newline$\pm$\,7.26   & 5.40\newline$\pm$\,6.02 \\
		\addlinespace[2pt]
		\rowcolor{gray!8}
		CMC-Seg &
		77.09\newline$\pm$\,31.79 & 87.53\newline$\pm$\,12.04 & 80.73\newline$\pm$\,25.28 &
		78.63\newline$\pm$\,29.91 & 81.20\newline$\pm$\,14.37 &
		4.90\newline$\pm$\,9.30   & 5.15\newline$\pm$\,9.06  & 5.65\newline$\pm$\,11.91 &
		4.39\newline$\pm$\,7.24   & 5.20\newline$\pm$\,5.83 \\
		\bottomrule[1.2pt]
	\end{tabularx}
	
	\vspace{3pt}
	\parbox{\textwidth}{\footnotesize
		Model (B) uses simple Mamba blocks; Model (C) uses plain skip connections; Model (D) uses simple Mamba blocks and plain skip connections. Higher Dice and lower HD95 are better ($\uparrow$/$\downarrow$).
		NETC, non-enhancing tumor core; SNFH, surrounding non-enhancing FLAIR hyperintensity; ET, enhancing tumor; RC, resection cavity.}
\end{table*}

\begin{figure*}[!t]
	\centering
	\includegraphics[width=\textwidth,height=0.88\textheight,keepaspectratio]{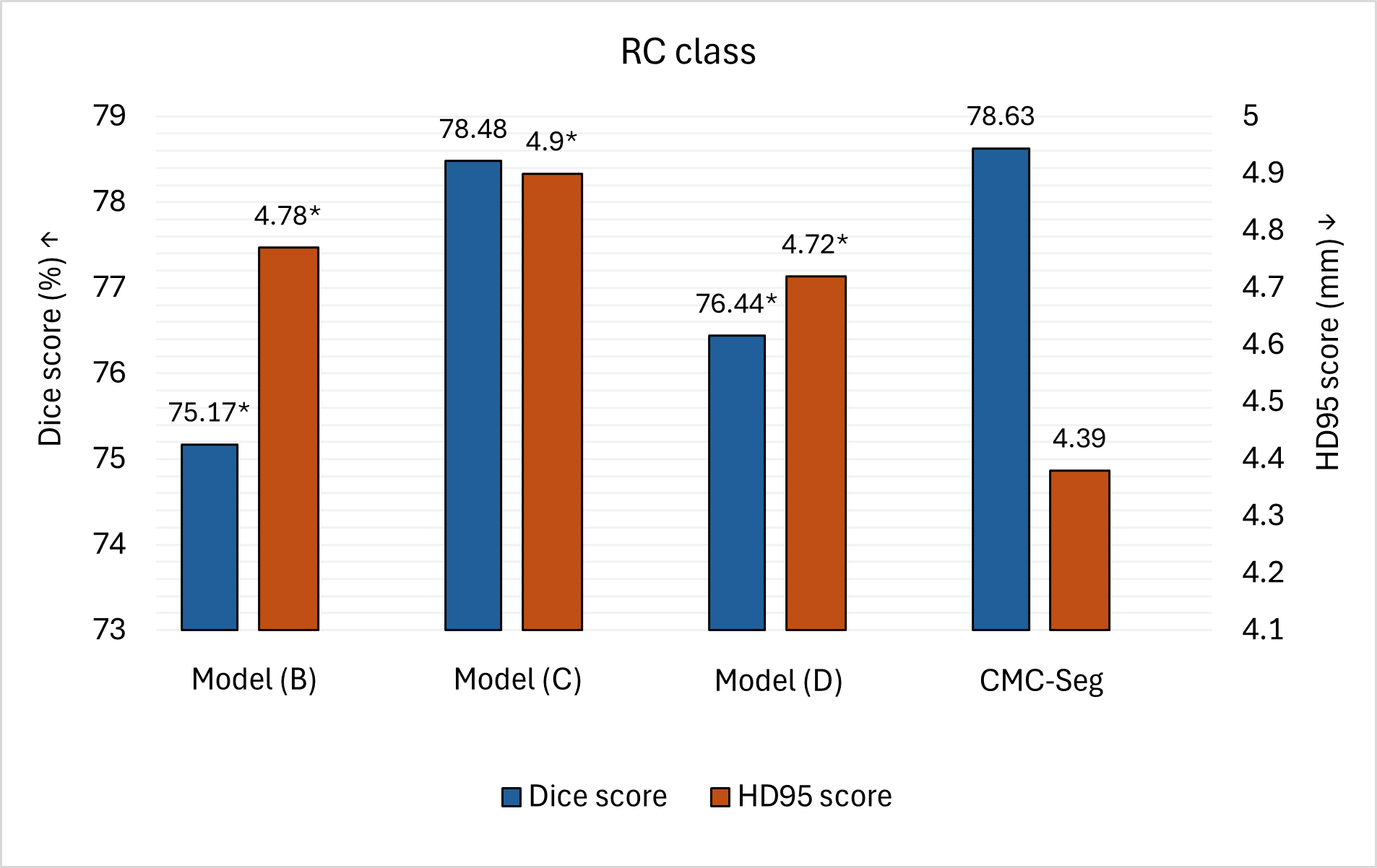}
	\caption{Effectiveness of the multi-directional Mamba and cross-attention strategy on the RC class.  Model (B) uses simple Mamba blocks; Model (C) uses plain skip connections; Model (D) uses simple Mamba blocks and plain skip connections. The combination of multi-directional Mamba and cross-attention achieved the highest Dice score and the lowest HD95 score for the RC class. Asterisks indicate statistically significant differences.}
	\label{FIG:6}
\end{figure*}

The results show a performance drop compared to our full model, with the mean Dice score falling to 79.93\% $\pm$ 15.04\% and the mean HD95 increasing to 5.28 $\pm$ 6.08 mm. Also, per-class analysis shows that the removal of the MDM block causes a significant decrease in the Dice score as well as a significant increase in the HD95 score for the RC class (Fig.~6). This finding shows that simply applying a 1D sequence model to volumetric data is inherently flawed, as it disregards the fundamental 3D spatial structure. By processing the volume along all three orthogonal axes, our MDM block preserves this crucial spatial context and creates a far more powerful representation of 3D anatomy, which is vital for accurate segmentation. 

In the second step, we validated our cross-attention strategy. We removed our proposed cross-attention module from the skip connections, using direct concatenation of the encoder feature map with the decoder (Model C). This degraded both the mean Dice score and the mean HD95 score compared to our full model. Also, per-class analysis shows that the removal of the cross-attention causes a significant increase in the HD95 score for the RC class. This insight underscores the importance of cross-attention. Simple concatenations are static operations; they combine features uniformly and in a content-agnostic way. In contrast, our cross-attention strategy is dynamic and content-aware, providing high-level global context from the bottleneck layer to selectively inform feature reuse at every encoder resolution before fusion via residual gated normalization. This learned, adaptive control over feature fusion is a crucial factor behind the superior performance and robustness of the CMC-Seg.

Finally, the model variant without both multi-direction Mamba and cross-attention (Model D) yields the worst mean HD95 score among the variants. In addition, CMC-Seg and Model (B) differ significantly in Dice and HD95 scores for the RC class. Although Model (D)'s mean Dice score was higher than Model (B), this shows that the cross-attention strategy improved the mean Dice score only when combined with the MDM block. In conclusion, the ablation studies systematically demonstrated that each component of CMC-Seg, including the MDM block and the cross-attention strategy, is well justified by its performance.

\subsection{Explainability experiment}
This subsection evaluates the transparency and clinical interpretability of the proposed 3D segmentation model by assessing the Grad-CAM explanation maps. A fundamental challenge in medical image segmentation is multi-class explainability. Unlike classification models that produce a single class-based heatmap, this research extends Grad-CAM to all segmented classes, providing a specific heatmap for each tumor class. 

We generated saliency maps from three model layers: the decoder's third and last residual blocks and the bottleneck layer. The heatmap image represents the MRI areas that the model emphasizes throughout its decision-making. This heatmap is created using the model's activations and is overlaid on the original image. It generally utilizes a color format of red, yellow, and blue to denote levels of high, medium, and low importance, respectively. Warm colors, like red and yellow, denote the zones where the model computes the highest value in recognizing the tumor. The tumor zone is marked in red or yellow, suggesting that these regions contributed most to the relevant class logit. The CAM heatmap utilizes color schemes to represent the model's emphasis on the NETC, SNFH, ET, and RC zones during glioma segmentation. The closer an area is to red, the more important the model considers it for segmenting the corresponding tumor class. 

\begin{figure*}
	\centering
	\includegraphics[width=\textwidth]{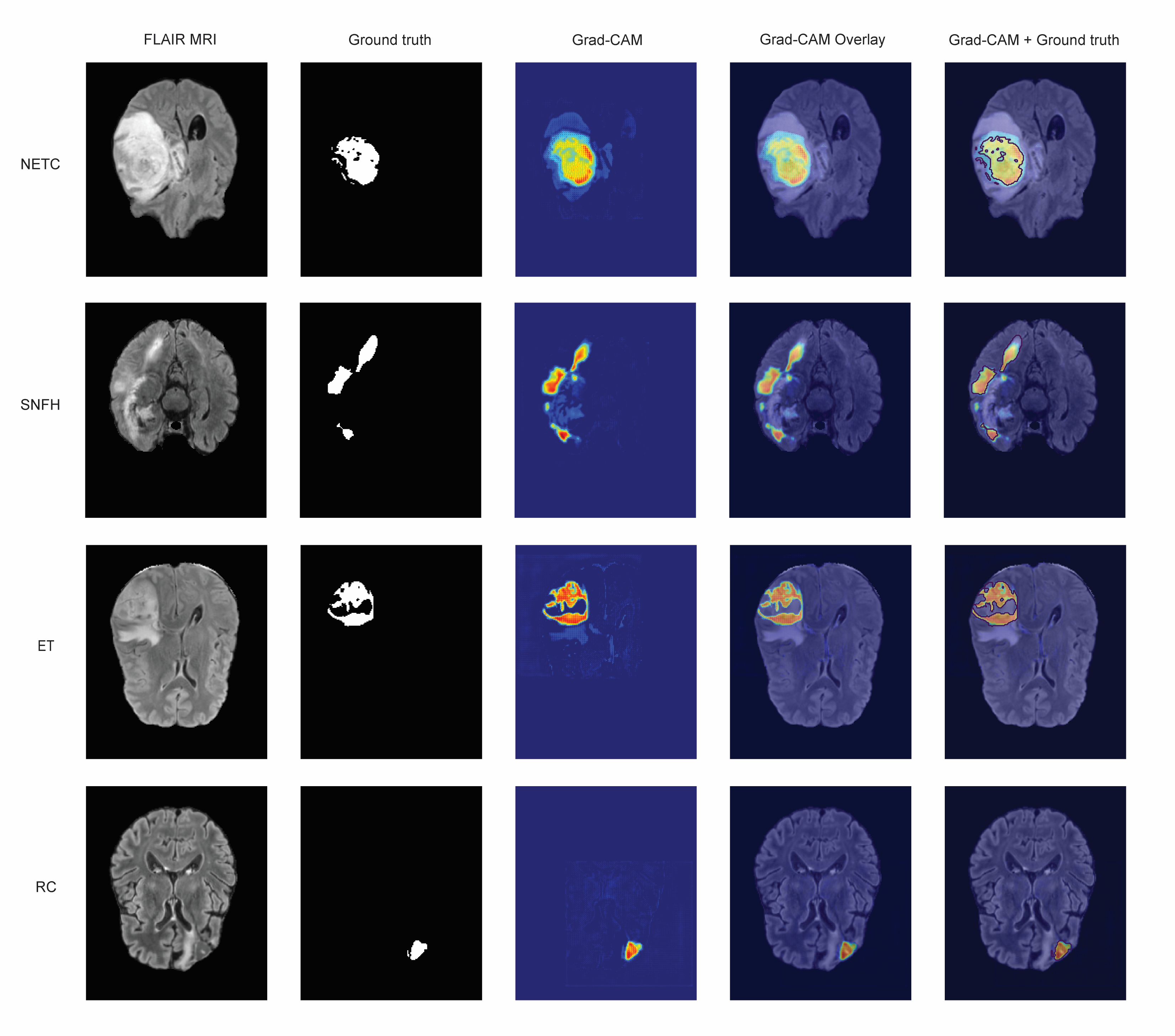}
	\caption{Visualization of per-class internal feature activation maps (heatmaps) from the last residual block of CMC-Seg.}
	\label{FIG:7}
\end{figure*}

\begin{figure*}
	\centering
	\includegraphics[width=\textwidth]{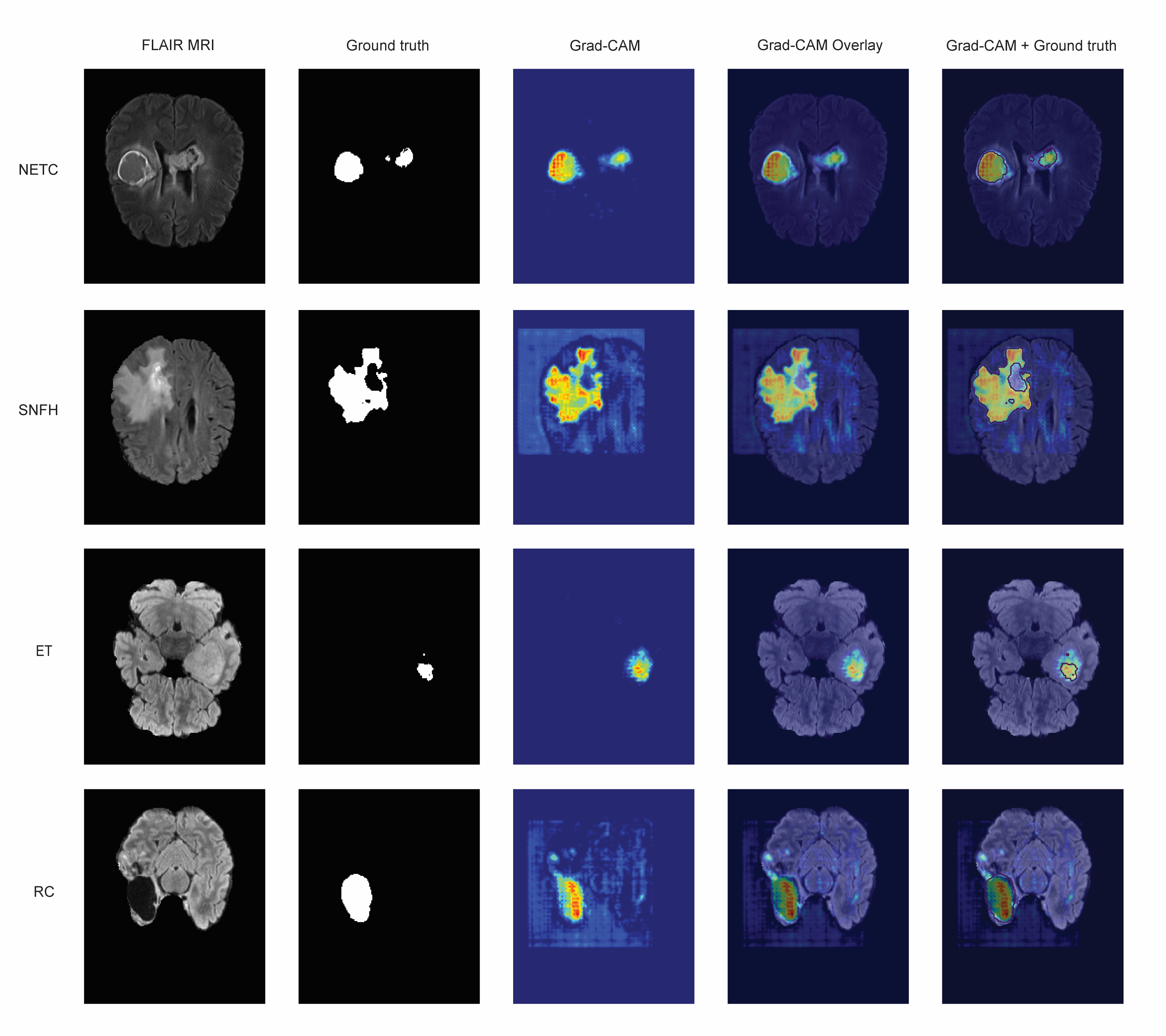}
	\caption{Visualization of per-class internal feature activation maps (heatmaps) from the third decoder residual block of CMC-Seg.}
	\label{FIG:8}
\end{figure*}

\begin{figure*}
	\centering
	\includegraphics[width=\textwidth]{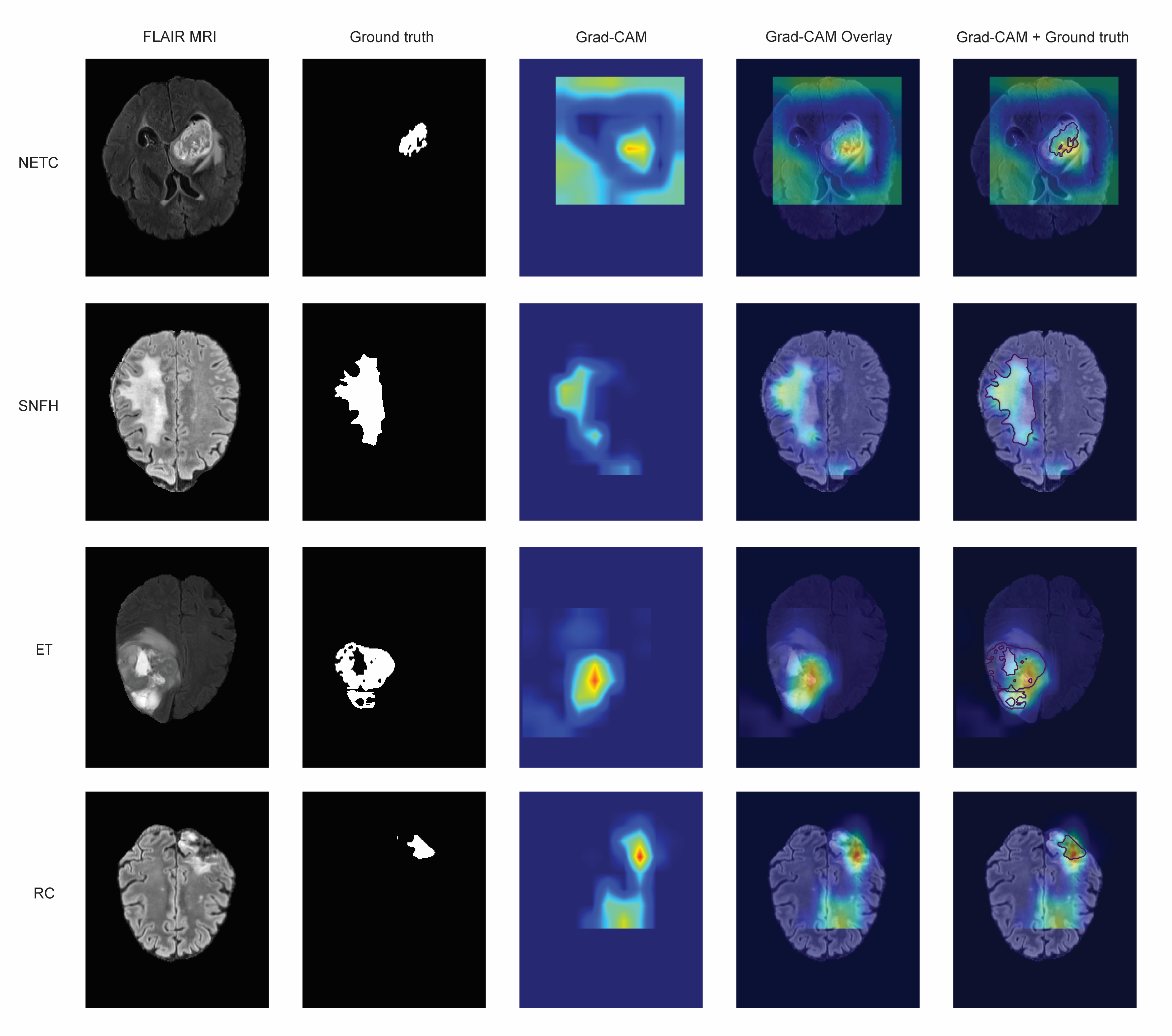}
	\caption{Visualization of per-class internal feature activation maps (heatmaps) from the bottleneck layer of CMC-Seg.}
	\label{FIG:9}
\end{figure*}

To provide a more intuitive understanding of our XAI results, we visualize the heatmaps of tumor classes from various layers in Fig.~7 through~9. Grad-CAM helps clinicians and end-users understand which parts of the MRI image the model considers most important for tumor segmentation.

Visualizing per-class heatmaps enhances transparency, enabling healthcare professionals to verify the AI-driven segmentation methodology used for diagnosis and patient monitoring. These results highlight the effective use of XAI in deep learning architectures, providing trustworthy models with potential clinical transparency and utility in medical imaging.

\section{Discussion}
In this study, we proposed a novel vision Mamba-CNN model enhanced with cross-attention (CMC-Seg) for 3D medical image segmentation. Our rigorous evaluations show its competitive computational efficiency and performance. Here, we go beyond the numerical results to discuss the broader implications of our findings, the intrinsic constraints of the methodology, and promising avenues for further investigation. 

The most important aspect of CMC-Seg's performance is how well it combines local feature extraction and global context modeling by developing an architecture that efficiently utilizes both, rather than choosing one over the other. As a fine-detail expert, the CNN component excels at capturing the subtle textural changes that define tumor margins. In contrast, the MDM-Mamba block serves as a contextual expert, capturing spatial interactions between a tumor and surrounding tissues as well as global anatomy. In the final stage, the cross-attention strategy refines the encoder's raw features by attending to bottleneck features before fusing them with decoder features. 

In addition, our cascade pipeline is the main factor that enables us to train our model and other benchmark models within such a low-compute-budget system and achieve competitive results on a class-imbalanced 3D dataset in only 100 epochs. Because deep learning models are opaque, we use Grad-CAM as an XAI method on several CMC-Seg layers to produce heatmaps. Analysis of these heatmaps shows that, by producing per-class heatmaps, the model improves transparency by linking predictions to clinically recognized anatomical structures and providing clinicians with a visual justification for the segmentation results. 

While we benchmarked against widely used models like 3D U-Net and SwinUNETR, our benchmark had limitations. For several reasons, including computational resource and storage limitations, we couldn't benchmark our model against nnU-Net\cite{isensee2021nnu}, U-Mamba\cite{ma2024u} and foundation models\cite{wang2025sam}. In addition, our benchmarking aimed to compare model architectures, not pipelines and frameworks. nnU-Net and U-Mamba are self-configuring pipelines with their own preprocessing, training, and postprocessing configurations. Therefore, it would be interesting to benchmark our cascaded pipeline against other pipelines to see its performance and efficacy. Additionally, cross-validation with six benchmark baselines and three ablation models was impractical due to computational constraints.

There is an array of aspects in the proposed framework that could potentially be improved and expanded upon in future work. The pipeline through ensemble methods and deep supervision can enhance robustness. Additionally, integrating novel architectures such as Gated Delta Networks\cite{yang2025gated} and Forgetting Transformers\cite{lin2025forgetting} holds promise for further accuracy gains. Including a larger and more diverse dataset is essential to improving generalizability. Future studies will focus on developing lighter models, handling missing MRI modalities, and extending the framework to support multi-tasking such as classification and prognosis prediction.

\section{Conclusion}
This work has proposed a cascade framework with CMC-Seg for pre- and post-treatment glioma segmentation, improving accuracy and computational efficiency. The two main components of CMC-Seg were multi-directional Mamba and cross-attention, and ablation experiments confirmed that both improved segmentation results. Our extensive evaluation showed competitive performance against well-established baselines, yielding the highest mean Dice score and the best mean Dice score-to-GFLOPs ratio. The explainability experiment results support that CMC-Seg is interpretable and clinically meaningful. These results establish the proposed framework as a potential and effective method for clinical implementation, especially in glioma diagnosis and treatment planning. Furthermore, accurate tumor segmentation provides a robust basis for further investigations into glioma classification and grading.

\section*{CRediT authorship contribution statement}
\textbf{Siavash Taremi:} Conceptualization, Methodology, Software, Data curation, Formal analysis, Investigation, Visualization, Writing -- original draft, Writing -- review \& editing. \textbf{Mahmoud Heidari:} Conceptualization, Methodology, Data curation, Formal analysis, Investigation, Writing -- review \& editing. \textbf{Seyed Mohammad Hossein Hashemi:} Conceptualization, Methodology, Software, Investigation, Writing -- review \& editing. \textbf{Mohsen Hooshmand:} Conceptualization, Methodology, Supervision, Project administration, Writing -- review \& editing.

\section*{Declaration of competing interests}
The authors declare that they have no known competing financial interests or personal relationships that could have appeared to influence the work reported in this paper.

\section*{Ethics approval}
Not applicable.

\section*{Funding}
This research did not receive any specific grant from funding agencies in the public, commercial, or not-for-profit sectors.

\section*{Data availability}
The BraTS 2025 dataset analyzed during the current study is available at: \url{https://www.synapse.org/Synapse:syn64153130/wiki/630130}

\bibliographystyle{unsrtnat}
\bibliography{references}  






\end{document}